\pdfoutput=1

\documentclass[11pt]{article}

\usepackage{acl}

\usepackage{times}
\usepackage{latexsym}
\usepackage{float}
\usepackage[T1]{fontenc}

\usepackage[utf8]{inputenc}

\usepackage{microtype}

\usepackage{inconsolata}

\usepackage{newtxmath}
\usepackage{graphicx}
\usepackage{makecell}
\usepackage{booktabs}
\usepackage{multirow}
\usepackage{hyperref}

\title{Mitigate Negative Transfer 
 with Similarity Heuristic Lifelong Prompt Tuning}

\author{
  Chenyuan Wu\textsuperscript{1},
  Gangwei Jiang\textsuperscript{1},
  Defu Lian\textsuperscript{1}\thanks{\hspace{1mm} Corresponding author.} \\
  \textsuperscript{1}University of Science and Technology of China\\
  \texttt{\{wuchenyuan, gwjiang\}@mail.ustc.edu.cn, liandefu@ustc.edu.cn}
  }

\begin{document}
\maketitle
\begin{abstract}
Lifelong prompt tuning has significantly advanced parameter-efficient lifelong learning with its efficiency and minimal storage demands on various tasks.
Our empirical studies, however, highlights certain transferability constraints in the current methodologies: a universal algorithm that guarantees consistent positive transfer across all tasks is currently unattainable, especially when dealing dissimilar tasks that may engender negative transfer.
Identifying the misalignment between algorithm selection and task specificity as the primary cause of negative transfer, we present the Similarity Heuristic Lifelong Prompt Tuning (SHLPT) framework. This innovative strategy partitions tasks into two distinct subsets by harnessing a learnable similarity metric, thereby facilitating fruitful transfer from tasks regardless of their similarity or dissimilarity. Additionally, SHLPT incorporates a parameter pool to combat catastrophic forgetting effectively. Our experiments shows that SHLPT outperforms state-of-the-art techniques in lifelong learning benchmarks and demonstrates robustness against negative transfer in diverse task sequences.\footnote{Source code is available at \url{https://github.com/wcyno23/SHLPT}.}

\end{abstract}

\section{Introduction}

Drawing on the remarkable capacity of humans to amass new knowledge throughout their lifetime, lifelong learning (LL) systems aim to emulate this progressive learning trajectory by sequentially mastering various tasks, each contributing to the system's cumulative knowledge base. 
However, it is not trivial for deep learning models to achieve this ideal due to inherent challenges. 
These include the need to (1) avoid catastrophic forgetting - where the acquisition of new information can lead to the erosion of previously learned knowledge, and to (2) promote efficient knowledge transfer - where the model can leverage past learning experiences to aid in the understanding and performance of future tasks. Addressing these challenges is crucial for the development of LL systems that can adapt and grow in a manner akin to human learning.

\begin{table}
    \centering
    \resizebox{\columnwidth}{!}{
    \begin{tabular}{cccc}
    \toprule
    \makecell[c]{Task \\ Source $\rightarrow$ Target} & \makecell[c]{Prompt\\ tuning\\(w/o transfer)} &\makecell[c]{Continual\\ Initialization\\(w/ transfer)} &\makecell[c]{Progressive\\ Prompts\\(w/ transfer)}\\
    \midrule
    Yahoo $\rightarrow$ AG News& \multirow{3}{*}{$86.25\pm 1.75$}& $\textbf{86.83}\pm\textbf{2.24}$&$85.33\pm1.61$ \\
    DBpedia $\rightarrow$ AG News & &$83.92\pm2.98$ &$85.00\pm1.73$ \\
    Amazon $\rightarrow$ AG News & &$85.50\pm1.75$ &$86.17\pm0.95$ \\
    AG News $\rightarrow$ Yahoo &\multirow{3}{*}{$67.03\pm0.46$} &$66.43\pm1.53$ &$65.17\pm2.11$ \\
    DBpedia $\rightarrow$ Yahoo &&$\textbf{67.73}\pm\textbf{1.10}$ &$67.13\pm1.65$ \\
    Amazon $\rightarrow$ Yahoo & &$66.43\pm1.53$ &$65.77\pm3.33$ \\
    Yahoo $\rightarrow$ DBpedia &\multirow{3}{*}{$97.86\pm0.50$} &$97.57\pm0.91$ &$97.94\pm0.38$ \\
    AG News $\rightarrow$ DBpedia & &$\textbf{98.33}\pm\textbf{0.42}$ &$97.81\pm0.89$ \\
    Amazon $\rightarrow$ DBpedia & &$97.40\pm0.47$ &$97.76\pm0.51$ \\
    DBpedia $\rightarrow$ Amazon &\multirow{3}{*}{$47.53\pm3.95$} &$48.86\pm1.10$ &$48.67\pm3.70$ \\
    Yahoo $\rightarrow$ Amazon & &$43.73\pm2.41$&$49.00\pm3.89$ \\
    AG News $\rightarrow$ Amazon & &$\textbf{50.73}\pm\textbf{4.32}$ &$50.60\pm1.20$ \\
    \bottomrule
    \end{tabular}
    }
    \caption{Transfer learning results on  AGNews, Yahoo, DBpedia, Amazon. We use accuracy as the metric. Continual initialization refers to initializing the prompt for the target task with the fine-tuned prompt obtained from the source task. Progressive Prompts \citep{razdaibiedina2022progressive} refers to concatenating the prompts fine-tuned on the source task onto the randomly initialized prompt of the target task.}
    \label{table:1}
\end{table}

Recent advancements in lifelong learning of language model (LM) have integrated the concept of prompt tuning to enhance its capabilities. These approaches maintain the pre-trained model's parameters, while training a small set of additional prompts to adapt the model to various downstream tasks.
The efficiency and lightweight nature of prompt tuning align well with the demands of LL, sidestepping the need for the onerous storage of entire model versions for every new task. This technique facilitates the model to accumulate knowledge over time, adapt flexibly to new tasks, and recall how to perform on older tasks with the aid of task-specific prompts. 
Moreover, substantial effort is dedicated to the transfer of knowledge from past tasks. This includes methods such as prompt concatenation~\citep{razdaibiedina2022progressive}, parameter sharing \citep{wang2022dualprompt, wang2022learning}, and weighted summation \citep{smith2023coda, jiang-etal-2023-towards}, which are pivotal in ensuring that knowledge is effectively retained and utilized throughout the lifelong learning process.

Current transfer learning methods commonly presuppose that earlier tasks can positively impact succeeding tasks.~\citet{razdaibiedina2022progressive} leverages prompts from previous tasks in the process of learning new ones, and similarly,~\citet{zhu-etal-2022-continual} utilizes prompts from former tasks as a foundation to harness prior knowledge. Nevertheless, we suggest that when there is a considerable dissimilarity between tasks, these approaches don't consistently assure positive transfer; on occasions, they might even provoke negative transfer. This phenomenon is vividly illustrated by the empirical analysis in Table~\ref{table:1}, which shows the transfer efficiency among diverse tasks within different soft prompt learning. On the other hand, how to effectively utilize negative transfer during the lifelong learning remains an open question for research. In this paper, we will further explore strategies for customized transfer learning tailored to the characteristics of different tasks, aimed at achieving more efficient knowledge accumulation while mitigating the potential impacts of negative transfer.

To address the issue of inconsistent knowledge transfer, we introduce a new approach to lifelong prompt tuning, which we refer as to SHLPT (\textbf{S}imilarity \textbf{H}euristic \textbf{L}ifelong \textbf{P}rompt \textbf{T}uning). First, we construct a prompt pool for learned tasks, thereby reducing the risk of forgetting. Then, we segment our knowledge transfer module into three components: (1) assessing the similarity between the current task and previous tasks, (2) categorizing the previous tasks into similar and dissimilar subsets, and (3) applying different transfer algorithms to each subset accordingly.
For the initial step, we calculate an attention-weighted combination of past prompt embeddings and incorporate this into the current task's prompt. During optimization, the model assigns higher attention scores to tasks that are more beneficial. Then, we utilize this attention score as task similarity metric and split task set accordingly.
For tasks deemed similar, we integrate their parameters to provide the current task with an optimized starting point. Conversely, for tasks that are dissimilar, we introduce a variety of innovative regularization techniques aimed at guiding the pre-trained model towards accessing a broader range of knowledge. This nuanced approach allows the model to better adapt to each new task while preserving and effectively utilizing the knowledge from all previous learning experiences without negative transfer.

In this paper, we make several notable contributions:
\textbf{(1)} We address the novel challenge of mitigating negative transfer and facilitating knowledge transfer from dissimilar tasks in lifelong learning, which is particularly important in sequences of low-similarity tasks. \textbf{(2)} We present SHLPT, an innovative lifelong prompt tuning technique that reduces forgetting and enables knowledge transfer across tasks with varying degrees of similarity. Our extensive experiments demonstrate that SHLPT surpasses existing methods on benchmark datasets. \textbf{(3)} We introduce a challenging benchmark characterized by low task similarity, which typically results in increased negative transfer. Our approach exceeds the performance of previous state-of-the-art methods in this context.

\section{Related Work}

\noindent\textbf{Lifelong Parameter Efficient Tuning.} Parameter efficient tuning tunes a subset of parameters of pre-trained language models, and can largely reduce the computation costs and memory usage \citep{houlsby2019parameter, li-liang-2021-prefix, hu2021lora, lester-etal-2021-power, ben-zaken-etal-2022-bitfit}. For lifelong prompt tuning, LFPT5 \citep{qin2021lfpt5} uses distillation loss and a generative replay to learn a continuous prompt. Progressive Prompts \citep{razdaibiedina2022progressive} learns a new prompt for every task, and progressively concatenate it with previous prompts. LPT \citep{liang-etal-2023-prompts} employs a trainable binary mask on the overall prompt to selectively choose parameters for different tasks. L2P \citep{wang2022learning} initializes a prompt pool and selects a certain number of prompts from it for each task. Through this approach, parameters among different tasks can be shared and isolated simultaneously. Base on this parameter pool architecture, CODA-prompt \citep{smith2023coda} employs an attention-based prompt selection strategy, Diana \citep{dai-etal-2023-domain} and HiDe-Prompt \citep{NEURIPS2023_d9f8b5ab} decompose prompts into a hierarchical structure. In addition to lifelong learning on prompt tuning, O-LoRA \citep{wang-etal-2023-orthogonal} introduces Orthogonal regularization to LoRA, facilitate learning in mutually orthogonal subspaces for different tasks; CLASSIC \citep{ke-etal-2021-classic} adds task masks to the Adapter layer and utilizes contrastive loss to transfer knowledge between similar tasks.

\noindent\textbf{Similarity Heuristic Methods.}
Similarity heuristic lifelong learning methods use task similarity to identify which tasks can transfer knowledge and minimize interference from dissimilar tasks. CAT \citep {ke2020continual} compares the performance between transfer model and reference model to determine whether use the transfer model. B-CL \citep{ke-etal-2021-adapting} and CTR \citep{ke2021achieving} utilize capsule network and routing algorithm to cluster similar tasks along with their shareable features. CLASSIC \citep{ke-etal-2021-classic} creates views form the hidden space information of previous similar tasks, and uses the contrastive loss to help current tasks to learn shared knowledge.

\section{Preliminary}

\subsection{Lifelong Learning Setup}
In lifelong learning, the language model will be sequentially finetuned across a series of $n$ tasks, labeled $T_1,...,T_n$. These tasks could originate from various domains and types. The training objective is to minimize the expected loss of all learned tasks with no access to data from previous tasks. Let $D_1, D_2, ..., D_n$ be the set of datasets corresponding to tasks $T_1, T_2, ..., T_n$, respectively. Each dataset $D_i$ encompasses a collection of data $(X^{i,j}, y^{i,j})$, where $X^{i,j} = [x_1, x_2, ..., x_l ]$ is the input text with length $l$, and $y^{i,j}$ is the corresponding output. The model is trained sequentially on these datasets with loss function $L_i$ specific to each task.

In this paper, we categorize LL scenarios into two types based on task similarity. In the first type~\citep{de2019episodic, razdaibiedina2022progressive}, the task sequence $T_1,...,T_n$ exhibits a relatively high degree of similarity, meaning that the likelihood of any task $T_t$ being significantly dissimilar from previous tasks $T_i (i<t)$ is low. The second type represents the opposite situation. The latter is more likely to emerge during the early stages of a real-world system when it has not yet accumulated a substantial knowledge base. We suppose that dissimilar tasks sequences are more prone to negative transfer, and for the first time, we create a lifelong learning benchmark composed of tasks that are dissimilar and likely to induce negative transfer for research purposes.

\subsection{Prompt Tuning}
Prompt tuning, as introduced by~\citet{lester-etal-2021-power}, presents a resource-efficient methodology for adapting language models without necessitating extensive fine-tuning of the original model. This technique utilizes a small set of trainable parameters known as a prompt, denoted $P = [p_1, p_2, \ldots, p_{l_p}]$, which serves as a prefixed sequence to the input token embeddings $X_e = [e(x_1), e(x_2), \ldots, e(x_l)]$. Here, $l_p$ denotes the length of the prompt, $e$ symbolizes the embedding function, and each vector $p_i \in \mathbb{R}^{d}$. For simplification, we use $X$ to replace $X_e$ below.

In the process of prompt tuning, the model employs the combined sequence $[P; X] \in \mathbb{R}^{(l_p+l) \times d}$ as the input for the LM. The objective function for adapting to a specific downstream task is given by $\mathcal L_{Down}([P; X]) = - \log p\left(y \mid [P; X]\right)$, which seeks to maximize the likelihood of the correct output $y$ given the input embedding $X$ and the prompt vectors $P$. Notably, this fine-tuning procedure exclusively modifies the prompt parameters $P$, leaving the remaining weights of the LM fixed.

\subsection{Empirical Study of Negative Transfer in Prompt Tuning}
\noindent\textbf{Definition of Negative Transfer.} Transfer learning leverages data or knowledge from source tasks to enhance target task's learning performance. However, the effectiveness of transfer learning is not always guaranteed \citep{zhang2022survey}; the performance on the target task may even worsen after transfer learning, a situation that is referred to as negative transfer \citep{pan2009survey}. Let $\mathcal{S}$ be one or more source tasks, $\mathcal{T}$ a target task, $\epsilon_{\mathcal{T}}$ target task's test error, $A(\mathcal{S},\mathcal{T})$ a transfer learning algorithm between $\mathcal{S}$ and $\mathcal{T}$, $A(\emptyset,\mathcal{T})$ the same algorithm without source tasks' information. The test error reduction after transfer learning can be formulated as \begin{equation}\label{1} r_{\mathcal{T}}=\epsilon_{\mathcal{T}}(A(\emptyset,\mathcal{T})) - \epsilon_{\mathcal{T}}(A(\mathcal{S},\mathcal{T}))\end{equation}A positive test error reduction $r_{\mathcal{T}}$ indicates a successful transfer learning result, while a negative value indicates negative transfer.

\begin{figure}[t]
    \centering
    \includegraphics[width=0.8\linewidth, scale=1.0]{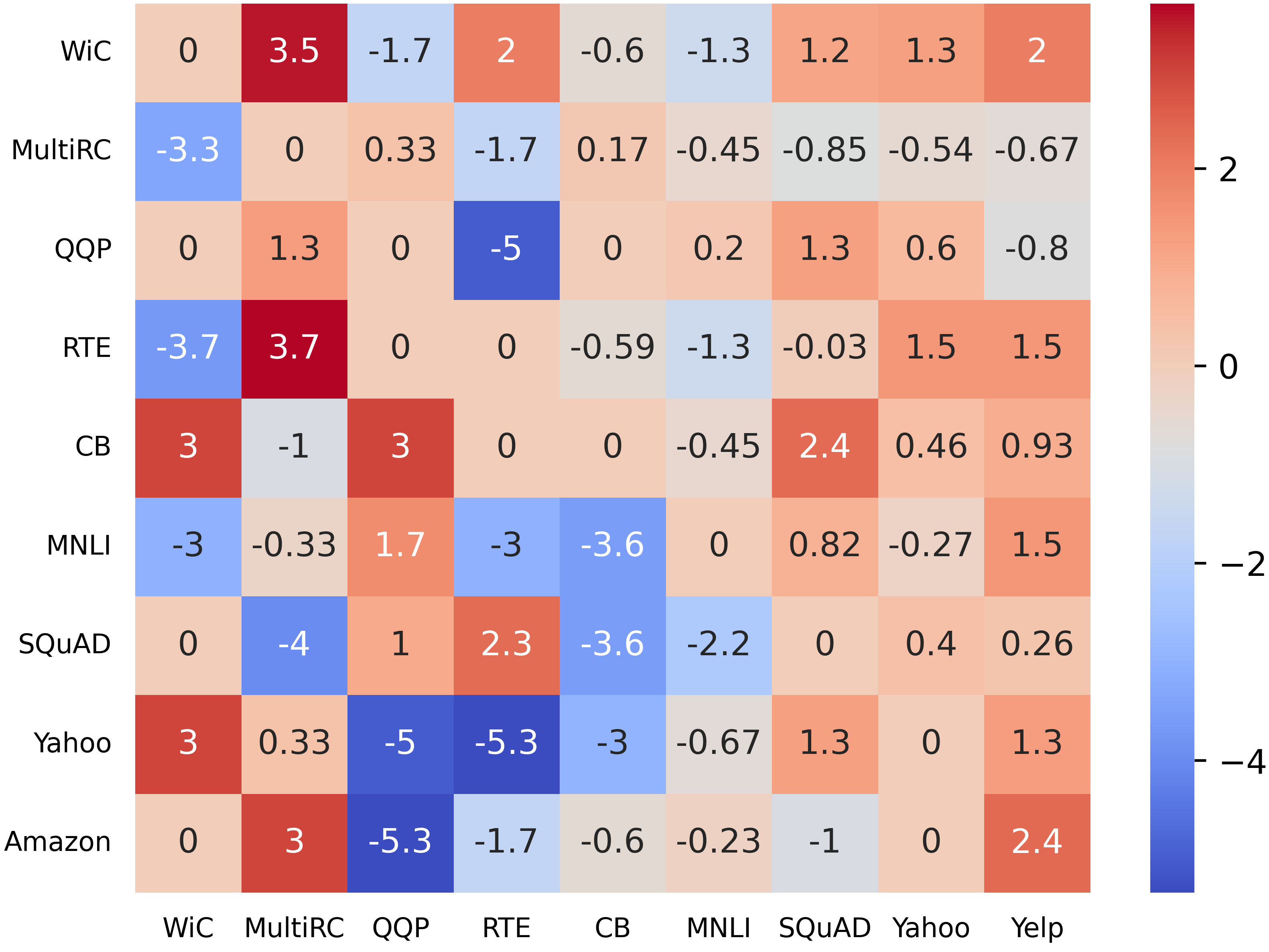}
    \caption{Test error reduction on the target tasks (column) after transferring from different source tasks (row). The negative transfer (indicated by cool colors) exists when use single transfer algorithm.} 
    \label{fig:1}
    \vspace{-1em}
\end{figure}

\begin{figure*}[tbp]
    \centering
    \includegraphics[width=\textwidth, scale=1.00]{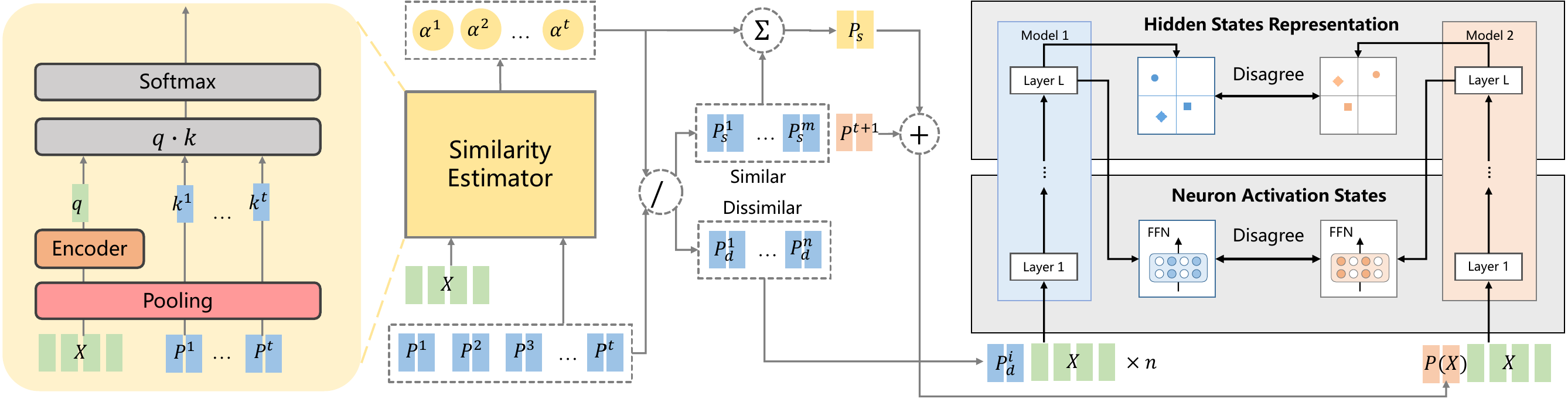}
    \caption{Illustration of our method SHLPT. The previous task prompts are partitioned based on an instance-wise similarity. Then, different transfer learning algorithm is applied on similar and dissimilar task scenarios. Similar tasks' prompts are composed and added to current task prompt. The current task's model behavior and representation are pushed away from those of dissimilar tasks. Only current task's prompt $P^{t+1}$ and encoder in similarity estimator are trainable.
 }
    \label{fig:2}
\end{figure*}

\noindent\textbf{Negative Transfer in Prompt Tuning.} \citet{vu-etal-2022-spot} shows many tasks can benefit each other under soft prompt transfer learning setting. In order to examine the negative transfer phenomenon in this setting, we conduct a empirical study on the transferability of soft prompt across a collection of language tasks (details on the tasks are available in Appendix~\ref{sec:appendix1}, while experimental specifics can be found in Appendix~\ref{sec:appendix3}). Here, a prompt learned on one source task is transferred and used as the initial prompt for a target task. As shown in Figure~\ref{fig:1}, the negative transfer is a common occurrence in the setting of soft prompt transfer, particularly when the source and target tasks are highly dissimilar – for instance, Yahoo$\rightarrow$RTE resulted in a $5.3\%$ drop in accuracy (from 78.67 to 73.33). When the source and target task pairs are similar, transfers tend to yield positive results. For instance, Amazon$\rightarrow$Yelp, both tasks involve sentiment analysis of reviews, leading to an $2.4\%$ increase in accuracy. 
Interestingly, positive transfer can sometimes occur even between dissimilar tasks (e.g., CB$\rightarrow$QQP). To better predict and avoid negative transfer, we design a similarity estimator that can discern transfer potential more effectively.

We further evaluated the few-shot (16 shot per class) performance of two task similarity agnostic transfer learning methods on four tasks from Standard CL Benchmark \citep{de2019episodic}: AGNews, Yahoo, DBpedia, Amazon. Results in Table~\ref{table:1} indicate that these two methods also cannot guarantee positive transfer among all tasks (e.g., DBpedia$\rightarrow$AG News, Amazon$\rightarrow$Yahoo). 

\noindent\textbf{Negative Transfer Benchmark.} Based on above results, we have developed a benchmark for lifelong learning research that consists of tasks prone to negative transfer
(Appendix~\ref{sec:appendix2}). This benchmark requires lifelong learning algorithms to effectively avoid negative transfer or further benefit from tasks involving negative transfer.

\noindent\textbf{Theoretical Analysis.} We further analyze the causes of negative transfer from theoretical bound for domain adaptation \citep{ben2010theory} .
\begin{equation}\label{2}
\small
\epsilon_{\mathcal{T}}(A(\mathcal{S},\mathcal{T})) \leq \epsilon_{\mathcal{S}}(A(\mathcal{S},\mathcal{T}))+\frac{1}{2} d_{\mathcal{H} \Delta \mathcal{H}}\left(D_{s}, D_{t}\right)+\lambda
\end{equation}
where $D$ is domain distribution, $d_{\mathcal{H} \Delta \mathcal{H}}\left(D_{s}, D_{t}\right)$ is the divergence between source domain $\mathcal{S}$ and target domain $\mathcal{T}$, $\lambda$ is a problem-specific constant. Based on the theoretical bound, negative transfer in LL can be attributed to the following reasons: (1) large error in the source tasks. This is caused by the few-shot setting or the learning difficulty of source tasks. (2) large distribution divergence. This is because the tasks are too dissimilar. (3) Unsuitable transfer learning algorithms. Transfer learning algorithms cannot adapt to all scenarios, an unsuitable algorithm may leads to negative transfer.

\section{Method}
To this end, we propose SHLPT (\textbf{S}imilarity \textbf{H}euristic \textbf{L}ifelong \textbf{P}rompt \textbf{T}uning), which intends to achieve the following goals for a robust lifelong learning system: (1) avoid forgetting previous knowledge, (2) transfer knowledge from both similar and dissimilar tasks, (3) reduce the probability of negative transfer during the transfer stage.

SHLPT is composed of the following three structures: (1) A prompt pool preserving previous task's prompts $\{P^1,P^2,...,P^t\}$ to avoid forgetting of past learning in a lifelong learning context. During the testing phase, identification of the task is necessary. (2) A task similarity estimator which can estimate the similarity between current tasks and previous tasks. We operate under the premise that "the more similar the tasks, the more effective the knowledge transfer." Based on estimated similarity, the collection of previous prompts is further categorized into two subsets: prompts from similar tasks ${P_s^1,P_s^2,...,P_s^m}$, and those from dissimilar tasks ${P_d^1,P_d^2,...,P_d^n}$. (3) Two transfer learning algorithms address transfer problems in different scenarios. One to disseminate shared knowledge extracted from similar tasks, and the other to differentiate core features from dissimilar tasks. Each structure is detailed below.

\subsection{Attention-based Similarity Estimator}
To tailor the transfer learning algorithm for appropriate tasks, we first conduct a similarity assessment and partitioning of the previous task. When training task $t$, the previous task's prompt embeddings are recorded as $\{P^1,P^2,...,P^{t-1}\}$. Rather than calculate similarity between different tasks, we calculate similarity between task $t$'s every instance and previous tasks, which is more robust to data variance \citep{wu-etal-2022-idpg}. 
Given that we lack access to data from past tasks, our similarity assessment relies on the prompts of previous tasks. 
Let $\alpha_i(X)$ to denote the instance-wise attention value for previous task $i$ given $X$. It is calculated by extracting features from three elements:  previous task prompts$\{P^1,P^2,...,P^{t-1}\}$, current task prompt $P^t$ and instance embedding $X$. 
The max pooling operation is first applied to $X\in\mathbb {R}^{l\times d}$ and $P^i\in\mathbb {R}^{l_p\times d}$, and transforms them into compact forms $\hat{X}, \hat{P^i}\in\mathbb {R}^{d}$, respectively.
$\hat{X}$ is then fed to a two-layer network to map it to the prompt space:
\begin{equation} \tilde{X} = LayerNorm(SiLU(\hat{X}W_{down})W_{up}), \end{equation}
where $W_{down}\in\mathbb {R}^{d\times r}$ and $ W_{up}\in\mathbb {R}^{r\times d} $ are trainable projection weights, SiLU \citep{elfwing2018sigmoid} is employed as an activation function and LayerNorm \citep{ba2016layer} is used to address the issue of gradient explosion, following \citet{asai-etal-2022-attempt}. Finally, the similarity $\alpha_i$ of previous task $i$ can be obtained through tempering softmax of inner product between $\tilde{X}$ and $\hat{P^i}$
\begin{equation} \alpha_i(X) = \frac{e^{\tilde{X}\hat{P^i}}/\tau_{sim}}{\sum_{j=1}^{t-1}e^{\tilde{X}\hat{P^j}}/\tau_{sim}} \end{equation} 
where $\tau_{sim}$ is a modifiable temperature parameter controlling the separation of similarity.
\citet{lian2020personalized} points out that the parameter $\tau$ in tempering softmax is an important factor controlling the output distribution. If $\tau$ is too large, the output is close to the uniform distribution. Conversely, if $\tau$ is too small, there will be a gradient vanishing problem. The temperature should be neither too large nor too small \citep{lian2020lightrec}.

Having obtained the task similarity between current instance and previous tasks, previous tasks can be easily partitioned based on it. We compare similarity value $\alpha_i$ with threshold $\delta$. 
If $\alpha_i > \delta$, the prompt corresponding to task $i$ is classified into similar task prompt set $\{P^{s_1},P^{s_2},...,P^{s_m}\}$, otherwise it will be moved into dissimilar set $\{P^{d_1},P^{d_2},...,P^{d_n}\}$. Here $s_i$ and $d_j$ denotes the original indicate of $i$-th similar task and $j$-th dissimilar task, respectively.
\begin{equation}
\left\{\begin{matrix} 
  P\to \{P^{s_1},P^{s_2},...,P^{s_m}\},& \text{if } \alpha > \delta   \\  
  P\to \{P^{d_1},P^{d_2},...,P^{d_n}\},& \text{if } \alpha \le  \delta  
\end{matrix}\right. 
\end{equation}
We use a predefined threshold to clearly partition the previous task into two sets, enabling the model to avoid negative transfer by applying the suitable transfer algorithm for different task scenario.

\subsection{Similar Task Transferring}
Towards previous tasks similar to current tasks, we consider transferring by prompt initialization perspective. Previous work has shown prompt tuning is sensitive to parameter initialization and usually suffer from slow convergence \citep{lester-etal-2021-power, vu-etal-2022-spot, wang2022multitask, shi2023dept}. Thus, we use a mixture of similar tasks' prompt embeddings to initialize prompt, and sum it with a newly allocated prompt $P^t$ for the current task. This method not only enhances model performance on current task, but also reduces the overall training time. The final prompt is derived as:
\begin{equation}
P(X)=\sum_{j=1}^{t-1}\tilde{\alpha_j}(X)P^j+P^t\label{eq:7}
\end{equation}
Here, only the current prompt $P^t$ and weights in similar estimator network are trainable. The prompt $P^j$ of previous tasks are fixed, thus preventing forgetting or backward regression on those tasks. 
The dissimilar tasks' prompts are explicitly excluded to reduce their interference. Mixture value $\tilde{\alpha_i}$ is obtained from following steps. First, per task similarity is input into a threshold function to let dissimilar tasks' similarity value to 0.
\begin{equation}
    \hat{\alpha_i} =
\begin{cases}
    \alpha_i, & \text{if } \alpha_i > \delta \\
    0, & \text{if }  \alpha_i \le  \delta
\end{cases}
\end{equation}
Then, a normalization operation is applied on it to let the overall sum equal to 1. 
At the end of current task training stage, all training samples' similarity $\alpha$ are averaged and used to calculate final prompt embedding $P$ of current task. Then $P$ is added to the prompt pool.
\subsection{Dissimilar Task Transferring}
Exclude dissimilar tasks from specific transfer learning algorithm can also mitigate negative transfer. Furthermore, we explore the use of an alternative transfer learning algorithm capable of leveraging dissimilar tasks to facilitate positive transfer effects. This is a novel problem that do not mentioned in previous lifelong learning researches. 

Rather than transfer knowledge from prompt parameters \citep{asai-etal-2022-attempt, wang2022multitask}, 
our approach leverages the knowledge embedded in a pre-trained model with selected prompts. Since higher layers of a pre-trained model often exhibit more task-specific behavior \citep{liu-etal-2019-linguistic}, we introduce two novel loss functions that based on language model's inner representation. These functions are designed to differentiate the behaviors of the current task from those of dissimilar tasks.

\subsubsection{Hidden States Contrastive Loss(HSC)}
Recognizing that dissimilar tasks may offer limited shareable knowledge with the current task, our approach emphasizes the divergence in the output representations of these tasks. we prepend the transferred current task's prompt $P(X)$, as well as the prompts from the dissimilar tasks $\{P^{d_1},P^{d_2},...,P^{d_n}\}$, and input the combined results into a pre-trained language model. 
We then calculate the last hidden states of the decoder, $h$ for the current task, and $h_1,...,h_n$ for the $n$ dissimilar tasks, all based on the same instance. The pairs $(h,h_1),...,(h,h_n)$ are treated as negative examples, while $(h,h)$ is treated as a positive example.  
The hidden states contrastive(HSC) loss is defined as 
\begin{equation}
    \scriptsize
    \mathcal{L}_{HSC} = 
    -\mathrm{log} \frac{\mathrm{exp}(\mathrm{cos}(h,h)/\tau_{hsc})}{\mathrm{exp}(\mathrm{cos}(h,h)/\tau_{hsc})+\sum_{k=1}^{n}\mathrm{exp}(\mathrm{cos}(h,h_k)/\tau_{hsc})} \\
\end{equation}
where $\mathrm{cos}$ refers to cosine similarity, $\tau_{hsc}$ refers to temperature parameter. The HSC loss is designed to diverge the hidden states representation between current task and dissimilar tasks for a same input.
\subsubsection{Activation States Contrastive Loss(ASC)}
Previous work has shown activation states of neurons in transformers' feed forward network are associated to specific behaviour (\citealp{geva-etal-2021-transformer}; \citealp{dai-etal-2022-knowledge}), and can be used to calculate task similarity \citep{su-etal-2022-transferability}. The feed forward network $\mathrm{FFN}$ can be formulated as
\begin{equation}
\mathrm{FFN}(h)=f(hW_i)W_o
\end{equation}
where $f$ is an activation function, $h$ is the hidden states and $W_i$,$W_o$ are parameter matrices.

We denote activation value $f(hW_i)$ as $s$. The activation states, computed as $sign(s)$, take the form of binary vectors where each element indicates the status of a particular neuron. 

Since the activation value $s$ contains more information, we use it for our implementation.
We then add a mask $m$ to filter out neurons activated by instance X.
\begin{equation}
m = 1 - sign(s_0), \hat{s} = s \odot m, 
\end{equation}
where $s_0$ is the activation value when no prompt is prepended, $\odot$ refers to element-wise multiply.
As previous probing experiments has shown higher layers' feature are more task-specific \citep{liu-etal-2019-linguistic}, we construct our activation contrastive loss based on last FFN layers' activation states from current task: $\hat{s}$ and dissimilar tasks: $\{\hat{s}_1,...,\hat{s}_n\}$
\begin{equation}
\scriptsize
    \mathcal{L}_{ASC} = 
    -\mathrm{log} \frac{\mathrm{exp}(\mathrm{cos}(\hat{s},\hat{s})/\tau_{asc})}{\mathrm{exp}(\mathrm{cos}(\hat{s},\hat{s})/\tau_{asc})+\sum_{k=1}^{n}\mathrm{exp}(\mathrm{cos}(\hat{s},\hat{s_k})/\tau_{asc})}
\end{equation}
where $\tau_{asc}$ refers to temperature parameter. 
The activation states contrastive loss aims to reduce the overlapping rate of activation states between current task and dissimilar tasks.

Finally, the overall loss is computed by
\begin{equation}
    \mathcal L=\mathcal L_{Down} + \lambda_1\mathcal L_{HSC}+ \lambda_2\mathcal L_{ASC}
\end{equation}
where $\mathcal L_{Down}$ is the standard prompt tuning loss on the downstream task. In addition, we use the hidden states and activation states at first position of the decoder output to compute contrastive losses.

\section{Experimental Setup}
\subsection{Datasets and Metrics}
We utilize three benchmarks to evaluate the model performance: 

\noindent\textbf{Standard CL Benchmark} is a widely used benchmark for lifelong language learning models' evaluation. It consists of four text classification datasets on different tasks or domains \citep{zhang2015character}: AGNews (topic classification), Yahoo (QA categorization), DBpedia (Wikipedia article classification), Amazon (sentiment analysis).

\noindent\textbf{Large Number of Tasks} consists of 15 classification tasks and is used to evaluate lifelong learning methods' performance on long sequences of tasks \citep{razdaibiedina2022progressive}. It includes four datasets from standard CL benchmark, four datasets from GLUE benchmark \citep{wang-etal-2018-glue}, five datasets from SuperGLUE benchmark \citep{wang2019superglue}, Yelp reviews \citep{zhang2015character} and IMDB reviews \citep{maas-etal-2011-learning}.

\noindent\textbf{Negative Transfer Benchmark} is a benchmark that we introduced to evaluate the robustness of a lifelong learning system under sequences of dissimilar/negative transfer tasks. We construct the benchmark in the following steps: First, we use initialization from the source task as the transfer algorithm and test which source/target task pair exhibits negative transfer in a collection of datasets: MNLI, QQP, RTE from GLUE benchmark \citep{wang-etal-2018-glue}, WiC, CB, COPA, BoolQ, MultiRC from SuperGLUE benchmark \citep{wang2019superglue},  SQuAD 2.0 \citep{rajpurkar-etal-2018-know}, Yahoo, Yelp and Amazon \citep{zhang2015character}. The experiment result is shown in Figure~\ref{fig:1} and Appendix~\ref{sec:appendix3}. Then, we construct three dissimilar task sequences, with the requirement that preceding tasks induce negative transfer on subsequent tasks. 

The detailed information regarding the task sequences are provided in the Appendix~\ref{sec:appendix2}. We use normalized F1 score \citep{mccann2018natural} for SQuAD, accuracy for other datasets. The task details and metrics are provided in the Appendix~\ref{sec:appendix1}.

\subsection{Baselines and Training Details}
We compare SHLPT with the following baselines, including recent SOTA methods. \textbf{Finetune}: continually finetunes the whole model parameters on sequences of tasks. \textbf{Online EWC} \citep{schwarz2018progress}: utilizes a regularization loss to constrain updates on crucial parameters associated with previous tasks. \textbf{ER} \citep{chaudhry2019tiny}: replays samples from previous tasks when training future tasks.
\textbf{Per-task Prompts}: trains each task with a separate prompt and keeps the remaining parameters fixed. This represents prompt tuning without transfer.
\textbf{L2P} \citep{wang2022learning}: maintains a prompt pool and selects prompts from it using an instance-wise query. \textbf{CODA-Prompt} \citep{smith2023coda}: implements instance-wise prompts through a weighted summation of prompts from the pool. \textbf{ProgPrompt} \citep{razdaibiedina2022progressive}: train a new prompt for each task and progressively concatenate it with prompts from old tasks.

To ensure a fair comparison with ProgPrompt and SHLPT, task identity is provided during test stage for L2P and CODA-prompt. While the task identity is not mandatory for these two methods, providing it can enhance their performance. The above methods are all implemented on original T5-large model. We report the mean results over three runs with different random seeds. The temperature $\tau_{sim}$ for similarity estimator is set to $2\times10^4$. The temperature $\tau$ for each contrastive loss is set to 1 while the weights $\lambda_1$ and $\lambda_2$ of the losses are set to 0.1 and 0.5 respectively. 
The similarity threshold $\delta$ is set to 0.06 for Standard CL Benchmark and Large Number of Tasks, and 1.5 for Negative Transfer Benchmark.
We include further analysis of the sensitivity of SHLPT to the similarity threshold, as well as additional training details, in Appendix~\ref{sec:appendix4}.

\begin{table*}[htbp]
\centering
\begin{small}
\begin{tabular}{@{}l|cccc|cccc@{}}
\toprule
\multicolumn{1}{l|}{\multirow{2}{*}{Method}} & \multicolumn{4}{c|}{Standard CL Benchmark}                                                                       & \multicolumn{4}{c}{Large Number of Tasks}                                                                       \\ \cmidrule(l){2-9} 
\multicolumn{1}{c|}{}                        & \multicolumn{1}{c}{Order1} & \multicolumn{1}{c}{Order2} & \multicolumn{1}{c|}{Order3} & \multicolumn{1}{c|}{Avg} & \multicolumn{1}{c}{Order4} & \multicolumn{1}{c}{Order5} & \multicolumn{1}{c|}{Order6} & \multicolumn{1}{c}{Avg} \\ \midrule
Finetune & 30.92 & 32.27 & \multicolumn{1}{c|}{35.94}& 33.04 & 13.40 & 11.58 &\multicolumn{1}{c|}{14.38} & 13.12   \\
Online EWC & 62.82 & 57.31& \multicolumn{1}{c|}{66.37} & 62.17 & 49.70& 49.82 & \multicolumn{1}{c|}{48.61} & 49.38 \\
ER & 64.63& 69.36& \multicolumn{1}{c|}{67.28}&67.09 & 60.34& 55.98& \multicolumn{1}{c|}{54.37} & 56.90\\
Per-task Prompts & 78.50 & 78.50 & \multicolumn{1}{c|}{78.50} & \multicolumn{1}{c|}{78.50} & 75.51 & 75.51 & \multicolumn{1}{c|}{75.51} & 75.51\\
L2P & 76.26& 75.63& \multicolumn{1}{c|}{74.97}& 75.62& 73.86& 73.67 & \multicolumn{1}{c|}{74.07} & 73.87   \\
CODA-Prompt & 77.01 & \textbf{80.16} & \multicolumn{1}{c|}{75.86}& 77.67 & 76.21& \textbf{77.02} & \multicolumn{1}{c|}{76.40} & 76.54 \\
ProgPrompt & 73.73 & 77.06 & \multicolumn{1}{c|}{78.13}& 76.31& 73.71& 72.52 &\multicolumn{1}{c|}{71.51} & 72.58  \\
SHLPT(ours) & \textbf{80.21} & 79.69 & \multicolumn{1}{c|}{\textbf{80.93}}& \textbf{80.28} & \textbf{77.62}  & 76.97 & \multicolumn{1}{c|}{\textbf{77.87}} & \textbf{77.49}\\
\bottomrule
\end{tabular}
\end{small}
\caption{Results on Standard CL Benchmark and Large Number of Tasks. We present the model's average accuracy after learning the last task. The standard deviations are provided in Appendix~\ref{sec:appendix7}.}
\label{table:2}
\end{table*}

\begin{table}[t]
\centering
\begin{small}
\begin{tabular}{@{}l|cccc@{}}
\toprule
\multicolumn{1}{l|}{\multirow{2}{*}{Method}} & \multicolumn{4}{c}{Negative Transfer Benchmark}\\ 
\cmidrule(l){2-5} 
\multicolumn{1}{c|}{} & \multicolumn{1}{c}{Seq1} & \multicolumn{1}{c}{Seq2} & \multicolumn{1}{c|}{Seq3} & \multicolumn{1}{c}{Avg} \\ 
\midrule
Finetune & 44.91 & 46.00 & \multicolumn{1}{c|}{22.95}& 37.95 \\
Online EWC & 70.68 & 59.30 &\multicolumn{1}{c|}{56.79} & 62.26 \\
ER & 58.18 & 66.26 & \multicolumn{1}{c|}{62.41}& 62.28 \\
Per-task Prompts & \textbf{83.45} & 81.65 & \multicolumn{1}{c|}{69.03} & 78.04\\
L2P & 82.60 & 80.59 & \multicolumn{1}{c|}{68.04}& 77.08 \\
CODA-Prompt & 82.66 & 80.08 & \multicolumn{1}{c|}{69.71} & 77.48\\
ProgPrompt & 78.79 & 80.00 & \multicolumn{1}{c|}{67.58} & 75.46\\
SHLPT(ours) & 83.37 & \textbf{82.47} & \multicolumn{1}{c|}{\textbf{70.16}}& \textbf{78.67}\\
\bottomrule
\end{tabular}
\end{small}
\caption{Results on Negative Transfer Benchmark. We report the average score after learning the last task. The standard deviations are provided in Appendix~\ref{sec:appendix7}.}
\vspace{-1.2em}
\label{table:3}
\end{table}

\section{Results and Analysis}
\subsection{Results on Existing Benchmark}
We first evaluate our method and other baselines on two existing benchmark: Standard CL Benchmark and Large Number of Tasks. Considering the task order may impact the results, we conduct experiments under three different task orders. Table~\ref{table:2} shows that our method SHLPT outperforms all baselines in these two existing benchmarks. SHLPT improves the recent SOTA with an increase of accuracy by $2.6\%$ on Standard CL Benchmark and $0.95\%$ on Large Number of Tasks. Since prompt-based methods can all avoid forgetting, we attribute the improvement of SHLPT to its better inter-task transfer effect.

In addition to the average accuracy metrics, we also provide the backward transfer scores and forward transfer scores in Appendix~\ref{sec:appendix8}. These scores are employed to evaluate the effectiveness of mitigating forgetting and negative transfer.

\subsection{Results on Negative Transfer Benchmark}
We also assess SHLPT using a challenge Negative Transfer Benchmark that we proposed. Our motivation is to test robustness to sequences of dissimilar tasks. When similarity is low, negative transfer is more likely to occur. This is a more realistic lifelong learning scenario, requiring the model to mitigate negative transfer and transfer knowledge from dissimilar tasks. We conduct experiments on three different sequences that exhibit negative transfer and the overall results are shown in Table~\ref{table:3}. SHLPT achieves an improvement of $1.2\%$ in average score over CODA-Prompt. These findings indicate that SHLPT exhibits greater robustness when confronted with sequences of tasks involving negative transfer. The backward transfer scores and forward transfer scores are reported in Appendix~\ref{sec:appendix8}.

\begin{table}
    \centering
    \resizebox{\columnwidth}{!}{
    \begin{small}
    \begin{tabular}{l|cccc|c}
    \toprule
     Model& Seq1 & Seq2 & Seq3 & Seq4 & Avg \\
    \midrule
    -ASE & 81.84 & 80.54 & 70.30 & 80.15 & 78.21\\
    -ASC & 83.22 & 82.22 & \textbf{70.92} & 79.69 & 79.01\\
    -HSC & \textbf{83.51} & 82.01 & 70.13 & 80.01 & 78.92\\
    -STT & 83.37 & \textbf{82.47} & 70.16 & 79.49 & 78.87\\
    -ASC,-HSC & 81.06 & 82.44 & 70.58 & 80.20 & 78.57\\
    \midrule
    \textbf{SHLPT}& 83.37 & \textbf{82.47} & 70.16 & \textbf{80.21} & \textbf{79.05}\\
    \bottomrule
    \end{tabular}
    \end{small}
    }
    \caption{Ablation experiment results. Seq1-3 refer to sequences from Negative Transfer Benchmark, Seq4 refers to the sequence from Standard CL Benchmark with Order1. The standard deviations are provided in Appendix~\ref{sec:appendix7}.}
    \label{table:5}
\end{table}

\begin{table*}[htbp]
    \centering
    \begin{small}
    \begin{tabular}{cccccc}
    \toprule
    \makecell[c]{Task \\ Source $\rightarrow$ Target} & \makecell[c]{Prompt\\ tuning\\(w/o transfer)} &\makecell[c]{Continual\\ Initialization\\(w/ transfer)} &\makecell[c]{Add HSC \\ Loss\\ (w/ transfer)} &\makecell[c]{Add ASC \\ Loss \\(w/ transfer)} &\makecell[c]{Add ASC\\\& HSC Loss\\(w/ transfer)}\\
    \midrule
    QQP $\rightarrow$ RTE  & \multirow{2}{*}{78.67}& \textcolor{blue}{73.67} & \textcolor{red}{83.00} & \textcolor{red}{79.00} & \textcolor{red}{82.33}\\
    Yahoo $\rightarrow$ RTE & & \textcolor{blue}{73.33} & \textcolor{red}{82.67} & \textcolor{red}{87.00} & \textcolor{red}{84.67}\\
    MNLI  $\rightarrow$ CB & \multirow{2}{*}{87.50}& \textcolor{blue}{83.93} & \textcolor{red}{91.67} &\textcolor{red}{91.07} & \textcolor{red}{91.67}\\
    SQuAD $\rightarrow$ CB   &  & \textcolor{blue}{83.93} & \textcolor{red}{92.26} & \textcolor{red}{89.88} & \textcolor{red}{89.29}\\
    COPA $\rightarrow$ QQP & \multirow{2}{*}{86.00}& \textcolor{blue}{81.00} &\textcolor{blue}{81.67} &\textcolor{blue}{85.67} & \textcolor{red}{86.67}\\
    Yahoo $\rightarrow$ QQP & & \textcolor{blue}{81.00} &\textcolor{blue}{85.33} & \textcolor{blue}{85.67} & \textcolor{blue}{85.33}\\
    COPA $\rightarrow$ MNLI & \multirow{2}{*}{88.67}& \textcolor{blue}{87.11} &\textcolor{red}{90.22} &\textcolor{red}{89.11} &\textcolor{blue}{88.22}\\
    SQuAD $\rightarrow$ MNLI  & & \textcolor{blue}{86.44} &\textcolor{red}{90.00} &\textcolor{red}{89.78} & \textcolor{red}{90.00}\\
    Average Accuracy & 85.21 & \textcolor{blue}{81.30} & \textcolor{red}{87.10} & \textcolor{red}{87.15} & \textcolor{red}{87.27}\\
    \bottomrule
    \end{tabular}
    \end{small}
     \caption{Different transfer learning algorithms' transfer learning results on negative transfer task pairs. Positive transfer are shown in red and negative transfer are shown in blue.}
    \label{table:4}
\end{table*}

\subsection{Ablation Studies}
We present the ablation results in Table~\ref{table:5}. "-ASE", "-ASC", "-HSC" and "-STT" denote models without Attention-based Similarity Estimator, without ASC loss, without HSC loss, without Similar Task Transferring, respectively. For "-ASE", we replace the similarity estimator with one that outputs the same similarity across all old tasks (for example, \{0.25, 0.25, 0.25, 0.25\}). “-STT” and SHLPT have same result in Seq1-3, because there is no similar task in Negative Transfer Benchmark. We observe substantial drops when eliminating “STT” in Seq4, indicating that it contributes to certain transfer benefits. Table~\ref{table:5} shows that every component is effective and the full SHLPT gives best results.

\subsection{Knowledge Transfer from Dissimilar Tasks}
We further investigate whether the proposed two contrastive losses can facilitate positive knowledge transfers from dissimilar tasks. The experiments are conducted on eight task pairs where negative transfer is observed when transferring through initialization. Table~\ref{table:4} demonstrates that in the majority of cases, these two losses have positive transfers on dissimilar tasks. Notably, the combination of both losses yields the best transfer effects ($+2.1\%$). This confirms SHLPT's effectiveness in transferring knowledge across diverse task sequences.

\subsection{Task Order Analysis}
As Table~\ref{table:2} shows, task order may affect the performance of SHLPT. For example, SHLPT's performance is relatively low on Order5. We further investigate the possible reasons for this order's impact on performance, as outlined below.

In Order5, SHLPT's performance on RTE is inferior compared to others, as shown in Table ~\ref{table:6}. The difference in RTE's previous tasks between three orders is that Order5 includes MultiRC, which is dissimilar to RTE in task type, and negative transfer (MultiRC $\rightarrow$ RTE) is observed in Table~\ref{table:app4}. This may indicate that SHLPT has poor transfer performance when transferring from MultiRC to RTE.
In conclusion, the task order can impact SHLPT's performance, as SHLPT has different transfer learning results based on different previous task sets.

\begin{table}[h]
    \centering
    \begin{small}
    \begin{tabular}{l|ccc}
    \toprule
     Task & RTE & QQP & Yahoo  \\
    \midrule
    Order4 & 84.17 & 87.17 & 74.43\\
    Order5 & 79.00 & 83.50 & 73.73\\
    Order6 & 85.00 & 85.33 & 76.70 \\
    \bottomrule
    \end{tabular}
    \end{small}
    \caption{Results of three tasks on which SHLPT performs poorly in Order5 compared to other orders.}
    \label{table:6}
\end{table}

The additional analysis about the training curves of task similarities and the impact of ASC loss on activation states is provided in Appendix~\ref{sec:appendix5} and Appendix ~\ref{sec:appendix6}.

\section{Conclusion}
In this paper, we introduce a novel method SHLPT for lifelong language learning. SHLPT aims to alleviate negative transfer by learning task similarity in one step and employing suitable transfer method for different old task scenarios (similar or dissimilar). Experimental results show SHLPT outperforms baseline methods via better knowledge transfer on two existing benchmarks and our custom Negative Transfer Benchmark.

\section*{Limitations}
Though our approach has achieve significant results, there are still some limitations. Firstly, SHLPT requires task identity at training and inference stage. The identification and mitigation of negative transfer in task agnostic lifelong learning setting remain undiscovered. Secondly, we do not evaluate SHLPT on multilingual tasks \citep{wang-etal-2020-negative} or other application tasks \citep{lian2014geomf, wang2019characterizing, NEURIPS2023_60f9118a}, and negative transfer may also occur in these scenario. Last, the scalability of SHLPT across different language models, especially LLM,
has not been thoroughly researched. We leave these for our future work.
\bibliography{anthology,custom}

\begin{thebibliography}{53}
\expandafter\ifx\csname natexlab\endcsname\relax\def\natexlab#1{#1}\fi

\bibitem[{Asai et~al.(2022)Asai, Salehi, Peters, and Hajishirzi}]{asai-etal-2022-attempt}
Akari Asai, Mohammadreza Salehi, Matthew Peters, and Hannaneh Hajishirzi. 2022.
\newblock \href {https://doi.org/10.18653/v1/2022.emnlp-main.446} {{ATTEMPT}: Parameter-efficient multi-task tuning via attentional mixtures of soft prompts}.
\newblock In \emph{Proceedings of the 2022 Conference on Empirical Methods in Natural Language Processing}, pages 6655--6672, Abu Dhabi, United Arab Emirates. Association for Computational Linguistics.

\bibitem[{Ba et~al.(2016)Ba, Kiros, and Hinton}]{ba2016layer}
Jimmy~Lei Ba, Jamie~Ryan Kiros, and Geoffrey~E Hinton. 2016.
\newblock Layer normalization.
\newblock \emph{arXiv preprint arXiv:1607.06450}.

\bibitem[{Ben-David et~al.(2010)Ben-David, Blitzer, Crammer, Kulesza, Pereira, and Vaughan}]{ben2010theory}
Shai Ben-David, John Blitzer, Koby Crammer, Alex Kulesza, Fernando Pereira, and Jennifer~Wortman Vaughan. 2010.
\newblock A theory of learning from different domains.
\newblock \emph{Machine learning}, 79:151--175.

\bibitem[{Ben~Zaken et~al.(2022)Ben~Zaken, Goldberg, and Ravfogel}]{ben-zaken-etal-2022-bitfit}
Elad Ben~Zaken, Yoav Goldberg, and Shauli Ravfogel. 2022.
\newblock \href {https://doi.org/10.18653/v1/2022.acl-short.1} {{B}it{F}it: Simple parameter-efficient fine-tuning for transformer-based masked language-models}.
\newblock In \emph{Proceedings of the 60th Annual Meeting of the Association for Computational Linguistics (Volume 2: Short Papers)}, pages 1--9, Dublin, Ireland. Association for Computational Linguistics.

\bibitem[{Chaudhry et~al.(2019)Chaudhry, Rohrbach, Elhoseiny, Ajanthan, Dokania, Torr, and Ranzato}]{chaudhry2019tiny}
Arslan Chaudhry, Marcus Rohrbach, Mohamed Elhoseiny, Thalaiyasingam Ajanthan, Puneet~K Dokania, Philip~HS Torr, and Marc'Aurelio Ranzato. 2019.
\newblock On tiny episodic memories in continual learning.
\newblock \emph{arXiv preprint arXiv:1902.10486}.

\bibitem[{Dai et~al.(2022)Dai, Dong, Hao, Sui, Chang, and Wei}]{dai-etal-2022-knowledge}
Damai Dai, Li~Dong, Yaru Hao, Zhifang Sui, Baobao Chang, and Furu Wei. 2022.
\newblock \href {https://doi.org/10.18653/v1/2022.acl-long.581} {Knowledge neurons in pretrained transformers}.
\newblock In \emph{Proceedings of the 60th Annual Meeting of the Association for Computational Linguistics (Volume 1: Long Papers)}, pages 8493--8502, Dublin, Ireland. Association for Computational Linguistics.

\bibitem[{Dai et~al.(2023)Dai, Lang, Zheng, Yu, Huang, and Li}]{dai-etal-2023-domain}
Yi~Dai, Hao Lang, Yinhe Zheng, Bowen Yu, Fei Huang, and Yongbin Li. 2023.
\newblock \href {https://doi.org/10.18653/v1/2023.findings-acl.361} {Domain incremental lifelong learning in an open world}.
\newblock In \emph{Findings of the Association for Computational Linguistics: ACL 2023}, pages 5844--5865, Toronto, Canada. Association for Computational Linguistics.

\bibitem[{de~Masson~D'Autume et~al.(2019)de~Masson~D'Autume, Ruder, Kong, and Yogatama}]{de2019episodic}
Cyprien de~Masson~D'Autume, Sebastian Ruder, Lingpeng Kong, and Dani Yogatama. 2019.
\newblock Episodic memory in lifelong language learning.
\newblock \emph{Advances in Neural Information Processing Systems}, 32.

\bibitem[{Elfwing et~al.(2018)Elfwing, Uchibe, and Doya}]{elfwing2018sigmoid}
Stefan Elfwing, Eiji Uchibe, and Kenji Doya. 2018.
\newblock Sigmoid-weighted linear units for neural network function approximation in reinforcement learning.
\newblock \emph{Neural networks}, 107:3--11.

\bibitem[{Geva et~al.(2021)Geva, Schuster, Berant, and Levy}]{geva-etal-2021-transformer}
Mor Geva, Roei Schuster, Jonathan Berant, and Omer Levy. 2021.
\newblock \href {https://doi.org/10.18653/v1/2021.emnlp-main.446} {Transformer feed-forward layers are key-value memories}.
\newblock In \emph{Proceedings of the 2021 Conference on Empirical Methods in Natural Language Processing}, pages 5484--5495, Online and Punta Cana, Dominican Republic. Association for Computational Linguistics.

\bibitem[{Houlsby et~al.(2019)Houlsby, Giurgiu, Jastrzebski, Morrone, De~Laroussilhe, Gesmundo, Attariyan, and Gelly}]{houlsby2019parameter}
Neil Houlsby, Andrei Giurgiu, Stanislaw Jastrzebski, Bruna Morrone, Quentin De~Laroussilhe, Andrea Gesmundo, Mona Attariyan, and Sylvain Gelly. 2019.
\newblock Parameter-efficient transfer learning for nlp.
\newblock In \emph{International Conference on Machine Learning}, pages 2790--2799. PMLR.

\bibitem[{Hu et~al.(2021)Hu, Wallis, Allen-Zhu, Li, Wang, Wang, Chen et~al.}]{hu2021lora}
Edward~J Hu, Phillip Wallis, Zeyuan Allen-Zhu, Yuanzhi Li, Shean Wang, Lu~Wang, Weizhu Chen, et~al. 2021.
\newblock Lora: Low-rank adaptation of large language models.
\newblock In \emph{International Conference on Learning Representations}.

\bibitem[{Jiang et~al.(2023{\natexlab{a}})Jiang, Jiang, Xue, Zhang, Zhou, Lian, and Wei}]{jiang-etal-2023-towards}
Gangwei Jiang, Caigao Jiang, Siqiao Xue, James Zhang, Jun Zhou, Defu Lian, and Ying Wei. 2023{\natexlab{a}}.
\newblock \href {https://doi.org/10.18653/v1/2023.findings-emnlp.808} {Towards anytime fine-tuning: Continually pre-trained language models with hypernetwork prompts}.
\newblock In \emph{Findings of the Association for Computational Linguistics: EMNLP 2023}, pages 12081--12095, Singapore. Association for Computational Linguistics.

\bibitem[{Jiang et~al.(2023{\natexlab{b}})Jiang, Chen, Pan, Wang, Liu, Jiang, and Long}]{NEURIPS2023_60f9118a}
Junguang Jiang, Baixu Chen, Junwei Pan, Ximei Wang, Dapeng Liu, Jie Jiang, and Mingsheng Long. 2023{\natexlab{b}}.
\newblock \href {https://proceedings.neurips.cc/paper_files/paper/2023/file/60f9118a849e8e9a0c67e2a36ad80ebf-Paper-Conference.pdf} {Forkmerge: Mitigating negative transfer in auxiliary-task learning}.
\newblock In \emph{Advances in Neural Information Processing Systems}, volume~36, pages 30367--30389. Curran Associates, Inc.

\bibitem[{Ke et~al.(2020)Ke, Liu, and Huang}]{ke2020continual}
Zixuan Ke, Bing Liu, and Xingchang Huang. 2020.
\newblock Continual learning of a mixed sequence of similar and dissimilar tasks.
\newblock \emph{Advances in Neural Information Processing Systems}, 33:18493--18504.

\bibitem[{Ke et~al.(2021{\natexlab{a}})Ke, Liu, Ma, Xu, and Shu}]{ke2021achieving}
Zixuan Ke, Bing Liu, Nianzu Ma, Hu~Xu, and Lei Shu. 2021{\natexlab{a}}.
\newblock Achieving forgetting prevention and knowledge transfer in continual learning.
\newblock \emph{Advances in Neural Information Processing Systems}, 34:22443--22456.

\bibitem[{Ke et~al.(2021{\natexlab{b}})Ke, Liu, Xu, and Shu}]{ke-etal-2021-classic}
Zixuan Ke, Bing Liu, Hu~Xu, and Lei Shu. 2021{\natexlab{b}}.
\newblock \href {https://doi.org/10.18653/v1/2021.emnlp-main.550} {{CLASSIC}: Continual and contrastive learning of aspect sentiment classification tasks}.
\newblock In \emph{Proceedings of the 2021 Conference on Empirical Methods in Natural Language Processing}, pages 6871--6883, Online and Punta Cana, Dominican Republic. Association for Computational Linguistics.

\bibitem[{Ke et~al.(2021{\natexlab{c}})Ke, Xu, and Liu}]{ke-etal-2021-adapting}
Zixuan Ke, Hu~Xu, and Bing Liu. 2021{\natexlab{c}}.
\newblock \href {https://doi.org/10.18653/v1/2021.naacl-main.378} {Adapting {BERT} for continual learning of a sequence of aspect sentiment classification tasks}.
\newblock In \emph{Proceedings of the 2021 Conference of the North American Chapter of the Association for Computational Linguistics: Human Language Technologies}, pages 4746--4755, Online. Association for Computational Linguistics.

\bibitem[{Lester et~al.(2021)Lester, Al-Rfou, and Constant}]{lester-etal-2021-power}
Brian Lester, Rami Al-Rfou, and Noah Constant. 2021.
\newblock \href {https://doi.org/10.18653/v1/2021.emnlp-main.243} {The power of scale for parameter-efficient prompt tuning}.
\newblock In \emph{Proceedings of the 2021 Conference on Empirical Methods in Natural Language Processing}, pages 3045--3059, Online and Punta Cana, Dominican Republic. Association for Computational Linguistics.

\bibitem[{Li and Liang(2021)}]{li-liang-2021-prefix}
Xiang~Lisa Li and Percy Liang. 2021.
\newblock \href {https://doi.org/10.18653/v1/2021.acl-long.353} {Prefix-tuning: Optimizing continuous prompts for generation}.
\newblock In \emph{Proceedings of the 59th Annual Meeting of the Association for Computational Linguistics and the 11th International Joint Conference on Natural Language Processing (Volume 1: Long Papers)}, pages 4582--4597, Online. Association for Computational Linguistics.

\bibitem[{Lian et~al.(2020{\natexlab{a}})Lian, Liu, and Chen}]{lian2020personalized}
Defu Lian, Qi~Liu, and Enhong Chen. 2020{\natexlab{a}}.
\newblock Personalized ranking with importance sampling.
\newblock In \emph{Proceedings of The Web Conference 2020}, pages 1093--1103.

\bibitem[{Lian et~al.(2020{\natexlab{b}})Lian, Wang, Liu, Lian, Chen, and Xie}]{lian2020lightrec}
Defu Lian, Haoyu Wang, Zheng Liu, Jianxun Lian, Enhong Chen, and Xing Xie. 2020{\natexlab{b}}.
\newblock Lightrec: A memory and search-efficient recommender system.
\newblock In \emph{Proceedings of The Web Conference 2020}, pages 695--705.

\bibitem[{Lian et~al.(2014)Lian, Zhao, Xie, Sun, Chen, and Rui}]{lian2014geomf}
Defu Lian, Cong Zhao, Xing Xie, Guangzhong Sun, Enhong Chen, and Yong Rui. 2014.
\newblock Geomf: joint geographical modeling and matrix factorization for point-of-interest recommendation.
\newblock In \emph{Proceedings of the 20th ACM SIGKDD international conference on Knowledge discovery and data mining}, pages 831--840.

\bibitem[{Liang et~al.(2023)Liang, Wei, Jie, Qian, Hao, and Han}]{liang-etal-2023-prompts}
Zujie Liang, Feng Wei, Yin Jie, Yuxi Qian, Zhenghong Hao, and Bing Han. 2023.
\newblock \href {https://doi.org/10.18653/v1/2023.acl-long.16} {Prompts can play lottery tickets well: Achieving lifelong information extraction via lottery prompt tuning}.
\newblock In \emph{Proceedings of the 61st Annual Meeting of the Association for Computational Linguistics (Volume 1: Long Papers)}, pages 277--292, Toronto, Canada. Association for Computational Linguistics.

\bibitem[{Liu et~al.(2019)Liu, Gardner, Belinkov, Peters, and Smith}]{liu-etal-2019-linguistic}
Nelson~F. Liu, Matt Gardner, Yonatan Belinkov, Matthew~E. Peters, and Noah~A. Smith. 2019.
\newblock \href {https://doi.org/10.18653/v1/N19-1112} {Linguistic knowledge and transferability of contextual representations}.
\newblock In \emph{Proceedings of the 2019 Conference of the North {A}merican Chapter of the Association for Computational Linguistics: Human Language Technologies, Volume 1 (Long and Short Papers)}, pages 1073--1094, Minneapolis, Minnesota. Association for Computational Linguistics.

\bibitem[{Loshchilov and Hutter(2017)}]{loshchilov2017decoupled}
Ilya Loshchilov and Frank Hutter. 2017.
\newblock Decoupled weight decay regularization.
\newblock \emph{arXiv preprint arXiv:1711.05101}.

\bibitem[{Maas et~al.(2011)Maas, Daly, Pham, Huang, Ng, and Potts}]{maas-etal-2011-learning}
Andrew~L. Maas, Raymond~E. Daly, Peter~T. Pham, Dan Huang, Andrew~Y. Ng, and Christopher Potts. 2011.
\newblock \href {https://aclanthology.org/P11-1015} {Learning word vectors for sentiment analysis}.
\newblock In \emph{Proceedings of the 49th Annual Meeting of the Association for Computational Linguistics: Human Language Technologies}, pages 142--150, Portland, Oregon, USA. Association for Computational Linguistics.

\bibitem[{McCann et~al.(2018)McCann, Keskar, Xiong, and Socher}]{mccann2018natural}
Bryan McCann, Nitish~Shirish Keskar, Caiming Xiong, and Richard Socher. 2018.
\newblock The natural language decathlon: Multitask learning as question answering.
\newblock \emph{arXiv preprint arXiv:1806.08730}.

\bibitem[{Pan and Yang(2009)}]{pan2009survey}
Sinno~Jialin Pan and Qiang Yang. 2009.
\newblock A survey on transfer learning.
\newblock \emph{IEEE Transactions on knowledge and data engineering}, 22(10):1345--1359.

\bibitem[{Paszke et~al.(2019)Paszke, Gross, Massa, Lerer, Bradbury, Chanan, Killeen, Lin, Gimelshein, Antiga et~al.}]{paszke2019pytorch}
Adam Paszke, Sam Gross, Francisco Massa, Adam Lerer, James Bradbury, Gregory Chanan, Trevor Killeen, Zeming Lin, Natalia Gimelshein, Luca Antiga, et~al. 2019.
\newblock Pytorch: An imperative style, high-performance deep learning library.
\newblock \emph{Advances in neural information processing systems}, 32.

\bibitem[{Qin and Joty(2021)}]{qin2021lfpt5}
Chengwei Qin and Shafiq Joty. 2021.
\newblock Lfpt5: A unified framework for lifelong few-shot language learning based on prompt tuning of t5.
\newblock In \emph{International Conference on Learning Representations}.

\bibitem[{Rajpurkar et~al.(2018)Rajpurkar, Jia, and Liang}]{rajpurkar-etal-2018-know}
Pranav Rajpurkar, Robin Jia, and Percy Liang. 2018.
\newblock \href {https://doi.org/10.18653/v1/P18-2124} {Know what you don{'}t know: Unanswerable questions for {SQ}u{AD}}.
\newblock In \emph{Proceedings of the 56th Annual Meeting of the Association for Computational Linguistics (Volume 2: Short Papers)}, pages 784--789, Melbourne, Australia. Association for Computational Linguistics.

\bibitem[{Razdaibiedina et~al.(2022)Razdaibiedina, Mao, Hou, Khabsa, Lewis, and Almahairi}]{razdaibiedina2022progressive}
Anastasia Razdaibiedina, Yuning Mao, Rui Hou, Madian Khabsa, Mike Lewis, and Amjad Almahairi. 2022.
\newblock Progressive prompts: Continual learning for language models.
\newblock In \emph{The Eleventh International Conference on Learning Representations}.

\bibitem[{Schwarz et~al.(2018)Schwarz, Czarnecki, Luketina, Grabska-Barwinska, Teh, Pascanu, and Hadsell}]{schwarz2018progress}
Jonathan Schwarz, Wojciech Czarnecki, Jelena Luketina, Agnieszka Grabska-Barwinska, Yee~Whye Teh, Razvan Pascanu, and Raia Hadsell. 2018.
\newblock Progress \& compress: A scalable framework for continual learning.
\newblock In \emph{International conference on machine learning}, pages 4528--4537. PMLR.

\bibitem[{Shi and Lipani(2023)}]{shi2023dept}
Zhengxiang Shi and Aldo Lipani. 2023.
\newblock Dept: Decomposed prompt tuning for parameter-efficient fine-tuning.
\newblock \emph{arXiv preprint arXiv:2309.05173}.

\bibitem[{Smith et~al.(2023)Smith, Karlinsky, Gutta, Cascante-Bonilla, Kim, Arbelle, Panda, Feris, and Kira}]{smith2023coda}
James~Seale Smith, Leonid Karlinsky, Vyshnavi Gutta, Paola Cascante-Bonilla, Donghyun Kim, Assaf Arbelle, Rameswar Panda, Rogerio Feris, and Zsolt Kira. 2023.
\newblock Coda-prompt: Continual decomposed attention-based prompting for rehearsal-free continual learning.
\newblock In \emph{Proceedings of the IEEE/CVF Conference on Computer Vision and Pattern Recognition}, pages 11909--11919.

\bibitem[{Su et~al.(2022)Su, Wang, Qin, Chan, Lin, Wang, Wen, Liu, Li, Li, Hou, Sun, and Zhou}]{su-etal-2022-transferability}
Yusheng Su, Xiaozhi Wang, Yujia Qin, Chi-Min Chan, Yankai Lin, Huadong Wang, Kaiyue Wen, Zhiyuan Liu, Peng Li, Juanzi Li, Lei Hou, Maosong Sun, and Jie Zhou. 2022.
\newblock \href {https://doi.org/10.18653/v1/2022.naacl-main.290} {On transferability of prompt tuning for natural language processing}.
\newblock In \emph{Proceedings of the 2022 Conference of the North American Chapter of the Association for Computational Linguistics: Human Language Technologies}, pages 3949--3969, Seattle, United States. Association for Computational Linguistics.

\bibitem[{Vu et~al.(2022)Vu, Lester, Constant, Al-Rfou{'}, and Cer}]{vu-etal-2022-spot}
Tu~Vu, Brian Lester, Noah Constant, Rami Al-Rfou{'}, and Daniel Cer. 2022.
\newblock \href {https://doi.org/10.18653/v1/2022.acl-long.346} {{SP}o{T}: Better frozen model adaptation through soft prompt transfer}.
\newblock In \emph{Proceedings of the 60th Annual Meeting of the Association for Computational Linguistics (Volume 1: Long Papers)}, pages 5039--5059, Dublin, Ireland. Association for Computational Linguistics.

\bibitem[{Wang et~al.(2019{\natexlab{a}})Wang, Pruksachatkun, Nangia, Singh, Michael, Hill, Levy, and Bowman}]{wang2019superglue}
Alex Wang, Yada Pruksachatkun, Nikita Nangia, Amanpreet Singh, Julian Michael, Felix Hill, Omer Levy, and Samuel Bowman. 2019{\natexlab{a}}.
\newblock Superglue: A stickier benchmark for general-purpose language understanding systems.
\newblock \emph{Advances in neural information processing systems}, 32.

\bibitem[{Wang et~al.(2018)Wang, Singh, Michael, Hill, Levy, and Bowman}]{wang-etal-2018-glue}
Alex Wang, Amanpreet Singh, Julian Michael, Felix Hill, Omer Levy, and Samuel Bowman. 2018.
\newblock \href {https://doi.org/10.18653/v1/W18-5446} {{GLUE}: A multi-task benchmark and analysis platform for natural language understanding}.
\newblock In \emph{Proceedings of the 2018 {EMNLP} Workshop {B}lackbox{NLP}: Analyzing and Interpreting Neural Networks for {NLP}}, pages 353--355, Brussels, Belgium. Association for Computational Linguistics.

\bibitem[{Wang et~al.(2023{\natexlab{a}})Wang, Xie, Zhang, Huang, Su, and Zhu}]{NEURIPS2023_d9f8b5ab}
Liyuan Wang, Jingyi Xie, Xingxing Zhang, Mingyi Huang, Hang Su, and Jun Zhu. 2023{\natexlab{a}}.
\newblock \href {https://proceedings.neurips.cc/paper_files/paper/2023/file/d9f8b5abc8e0926539ecbb492af7b2f1-Paper-Conference.pdf} {Hierarchical decomposition of prompt-based continual learning: Rethinking obscured sub-optimality}.
\newblock In \emph{Advances in Neural Information Processing Systems}, volume~36, pages 69054--69076. Curran Associates, Inc.

\bibitem[{Wang et~al.(2024)Wang, Zhang, Su, and Zhu}]{wang2024comprehensive}
Liyuan Wang, Xingxing Zhang, Hang Su, and Jun Zhu. 2024.
\newblock A comprehensive survey of continual learning: Theory, method and application.
\newblock \emph{IEEE Transactions on Pattern Analysis and Machine Intelligence}.

\bibitem[{Wang et~al.(2023{\natexlab{b}})Wang, Chen, Ge, Xia, Bao, Zheng, Zhang, Gui, and Huang}]{wang-etal-2023-orthogonal}
Xiao Wang, Tianze Chen, Qiming Ge, Han Xia, Rong Bao, Rui Zheng, Qi~Zhang, Tao Gui, and Xuanjing Huang. 2023{\natexlab{b}}.
\newblock \href {https://doi.org/10.18653/v1/2023.findings-emnlp.715} {Orthogonal subspace learning for language model continual learning}.
\newblock In \emph{Findings of the Association for Computational Linguistics: EMNLP 2023}, pages 10658--10671, Singapore. Association for Computational Linguistics.

\bibitem[{Wang et~al.(2022{\natexlab{a}})Wang, Panda, Karlinsky, Feris, Sun, and Kim}]{wang2022multitask}
Zhen Wang, Rameswar Panda, Leonid Karlinsky, Rogerio Feris, Huan Sun, and Yoon Kim. 2022{\natexlab{a}}.
\newblock Multitask prompt tuning enables parameter-efficient transfer learning.
\newblock In \emph{The Eleventh International Conference on Learning Representations}.

\bibitem[{Wang et~al.(2022{\natexlab{b}})Wang, Zhang, Ebrahimi, Sun, Zhang, Lee, Ren, Su, Perot, Dy et~al.}]{wang2022dualprompt}
Zifeng Wang, Zizhao Zhang, Sayna Ebrahimi, Ruoxi Sun, Han Zhang, Chen-Yu Lee, Xiaoqi Ren, Guolong Su, Vincent Perot, Jennifer Dy, et~al. 2022{\natexlab{b}}.
\newblock Dualprompt: Complementary prompting for rehearsal-free continual learning.
\newblock In \emph{European Conference on Computer Vision}, pages 631--648. Springer.

\bibitem[{Wang et~al.(2022{\natexlab{c}})Wang, Zhang, Lee, Zhang, Sun, Ren, Su, Perot, Dy, and Pfister}]{wang2022learning}
Zifeng Wang, Zizhao Zhang, Chen-Yu Lee, Han Zhang, Ruoxi Sun, Xiaoqi Ren, Guolong Su, Vincent Perot, Jennifer Dy, and Tomas Pfister. 2022{\natexlab{c}}.
\newblock Learning to prompt for continual learning.
\newblock In \emph{Proceedings of the IEEE/CVF Conference on Computer Vision and Pattern Recognition}, pages 139--149.

\bibitem[{Wang et~al.(2019{\natexlab{b}})Wang, Dai, P{\'o}czos, and Carbonell}]{wang2019characterizing}
Zirui Wang, Zihang Dai, Barnab{\'a}s P{\'o}czos, and Jaime Carbonell. 2019{\natexlab{b}}.
\newblock Characterizing and avoiding negative transfer.
\newblock In \emph{Proceedings of the IEEE/CVF conference on computer vision and pattern recognition}, pages 11293--11302.

\bibitem[{Wang et~al.(2020)Wang, Lipton, and Tsvetkov}]{wang-etal-2020-negative}
Zirui Wang, Zachary~C. Lipton, and Yulia Tsvetkov. 2020.
\newblock \href {https://doi.org/10.18653/v1/2020.emnlp-main.359} {On negative interference in multilingual models: Findings and a meta-learning treatment}.
\newblock In \emph{Proceedings of the 2020 Conference on Empirical Methods in Natural Language Processing (EMNLP)}, pages 4438--4450, Online. Association for Computational Linguistics.

\bibitem[{Wolf et~al.(2020)Wolf, Debut, Sanh, Chaumond, Delangue, Moi, Cistac, Rault, Louf, Funtowicz, Davison, Shleifer, von Platen, Ma, Jernite, Plu, Xu, Le~Scao, Gugger, Drame, Lhoest, and Rush}]{wolf-etal-2020-transformers}
Thomas Wolf, Lysandre Debut, Victor Sanh, Julien Chaumond, Clement Delangue, Anthony Moi, Pierric Cistac, Tim Rault, Remi Louf, Morgan Funtowicz, Joe Davison, Sam Shleifer, Patrick von Platen, Clara Ma, Yacine Jernite, Julien Plu, Canwen Xu, Teven Le~Scao, Sylvain Gugger, Mariama Drame, Quentin Lhoest, and Alexander Rush. 2020.
\newblock \href {https://doi.org/10.18653/v1/2020.emnlp-demos.6} {Transformers: State-of-the-art natural language processing}.
\newblock In \emph{Proceedings of the 2020 Conference on Empirical Methods in Natural Language Processing: System Demonstrations}, pages 38--45, Online. Association for Computational Linguistics.

\bibitem[{Wu et~al.(2022)Wu, Wang, Gu, Hou, Dong, Vydiswaran, and Ma}]{wu-etal-2022-idpg}
Zhuofeng Wu, Sinong Wang, Jiatao Gu, Rui Hou, Yuxiao Dong, V.G.Vinod Vydiswaran, and Hao Ma. 2022.
\newblock \href {https://doi.org/10.18653/v1/2022.naacl-main.403} {{IDPG}: An instance-dependent prompt generation method}.
\newblock In \emph{Proceedings of the 2022 Conference of the North American Chapter of the Association for Computational Linguistics: Human Language Technologies}, pages 5507--5521, Seattle, United States. Association for Computational Linguistics.

\bibitem[{Zhang et~al.(2022)Zhang, Deng, Zhang, and Wu}]{zhang2022survey}
Wen Zhang, Lingfei Deng, Lei Zhang, and Dongrui Wu. 2022.
\newblock A survey on negative transfer.
\newblock \emph{IEEE/CAA Journal of Automatica Sinica}, 10(2):305--329.

\bibitem[{Zhang et~al.(2015)Zhang, Zhao, and LeCun}]{zhang2015character}
Xiang Zhang, Junbo Zhao, and Yann LeCun. 2015.
\newblock Character-level convolutional networks for text classification.
\newblock \emph{Advances in neural information processing systems}, 28.

\bibitem[{Zhu et~al.(2022)Zhu, Li, Mi, Zhu, and Huang}]{zhu-etal-2022-continual}
Qi~Zhu, Bing Li, Fei Mi, Xiaoyan Zhu, and Minlie Huang. 2022.
\newblock \href {https://doi.org/10.18653/v1/2022.acl-long.80} {Continual prompt tuning for dialog state tracking}.
\newblock In \emph{Proceedings of the 60th Annual Meeting of the Association for Computational Linguistics (Volume 1: Long Papers)}, pages 1124--1137, Dublin, Ireland. Association for Computational Linguistics.

\end{thebibliography}

\clearpage

\appendix

\section{Dataset Details}
\label{sec:appendix1}
We present detailed information about datasets we used in Table~\ref{table:app1}. Following previous studies (\citealp{asai-etal-2022-attempt}; \citealp{razdaibiedina2022progressive}), we use datasets from \url{http://goo.gl/JyCnZq} for Standard CL Benchmark, while using HuggingFace datasets library (\url{https://github.com/huggingface/datasets}) for the remaining datasets.

\section{Task Sequence Details }
\label{sec:appendix2}
We use six differ orders of sequences in existing benchmark experiments, and the sequence information is presented in Table~\ref{table:app2}. For Negative Transfer Benchmark, we utilize three sequences composed of different datasets. The benchmark requires that preceding tasks induce negative transfer on subsequent tasks. Therefore, we do not alter the order as it may involve positive transfer. The sequence information is shown in Table~\ref{table:app3}.

\section{Soft Prompt Transfer Results}
\label{sec:appendix3}
The detailed results of our empirical study experiment are shown in Table~\ref{table:app4}. We perform transfer by initializing the target prompt from the source prompt. For classification tasks, we sample 100 samples per class to form the training set and validation set, and 100 samples per class to form the test set. For SQuAD dataset, we sample 400 samples to form the training set and validation set, and 400 samples to form the test set.

\section{Implementation Details}
\label{sec:appendix4}
We implement all methods with PyTorch \citep{paszke2019pytorch} and huggingface transformers \citep{wolf-etal-2020-transformers} library. All the experiments are run on eight NVIDIA 3090 GPUs. We set the max token length to 256 for all datasets. Following \citet{razdaibiedina2022progressive}, we use the available validation set as the test set and create validation set from the training set. If not specifically mentioned, for classification tasks, we sample 75 samples per class to form the training set and validation set, and 100 samples per class to form the test set, following \citet{qin2021lfpt5}. For SQuAD dataset, we sample 300 samples to form the training set and validation set, and 400 samples to form the test set. We use AdamW optimizer \citep{loshchilov2017decoupled} with a weight decay of 0.01 and a batch size of 8. All results are averaged over three runs with random seeds \{42, 142, 242\}.

For baselines that tuning all parameters (Finetune, online EWC, ER), we use the learning rate of $1\times 10^{-4}$. For prompt-based methods, we use the learning rate of $0.3$.

For SHLPT, we perform grid search on $\tau_{sim}$ within \{$2\times10^{4}, 2\times10^{5}, 2\times10^{6}, 2\times10^{7}$\}, $\tau_{hsc}$ and $\tau_{asc}$ within \{0.03, 0.5, 0.8, 1\}, and $\lambda_1$ and $\lambda_2$ within \{0.03, 0.1, 0.5, 0.8, 1.5\}. We set prompt token length to 150 and train the model with 80 epochs in Large Number of Tasks benchmark, while for others, we train 50 epochs. Early stopping mechanism is employed on all experiments. 
For similarity threshold $\delta$: as the Standard CL Benchmark and Large Number of Tasks contain some similar task pairs, we perform a grid search within the range of \{0.02, 0.04, ..., 0.20\}. We opt for a relatively low value range as we aim to ensure that no similar tasks are mistakenly categorized as dissimilar. Then we select 0.06 for these benchmarks. For Negative Transfer Benchmark, the previous task are all dissimilar from current task, so we search in a relatively high value range: \{0.50,...,1.50\}. And we select 1.5 for this benchmark.

Table~\ref{table:app5} reports the searching results of threshold on Standard CL Benchmark in Order1.
The results exhibits relatively minor changes as the threshold approaches 0. 
When the threshold increases, we notice that there is a certain performance decrease observed in the tasks towards the end of the task sequence. This is because the threshold value affects how and when SHLPT partitions previous tasks, thus further impacting the performance of the transfer algorithm. The tasks within the Standard CL Benchmark exhibit relatively high similarity, hence a lower threshold can prevent similar tasks from being partitioned as dissimilar.

\section{Similarity Training Curve}
\label{sec:appendix5}
\begin{figure}[htbp]
    \centering
    \includegraphics[width=0.68\linewidth, scale=1.00]{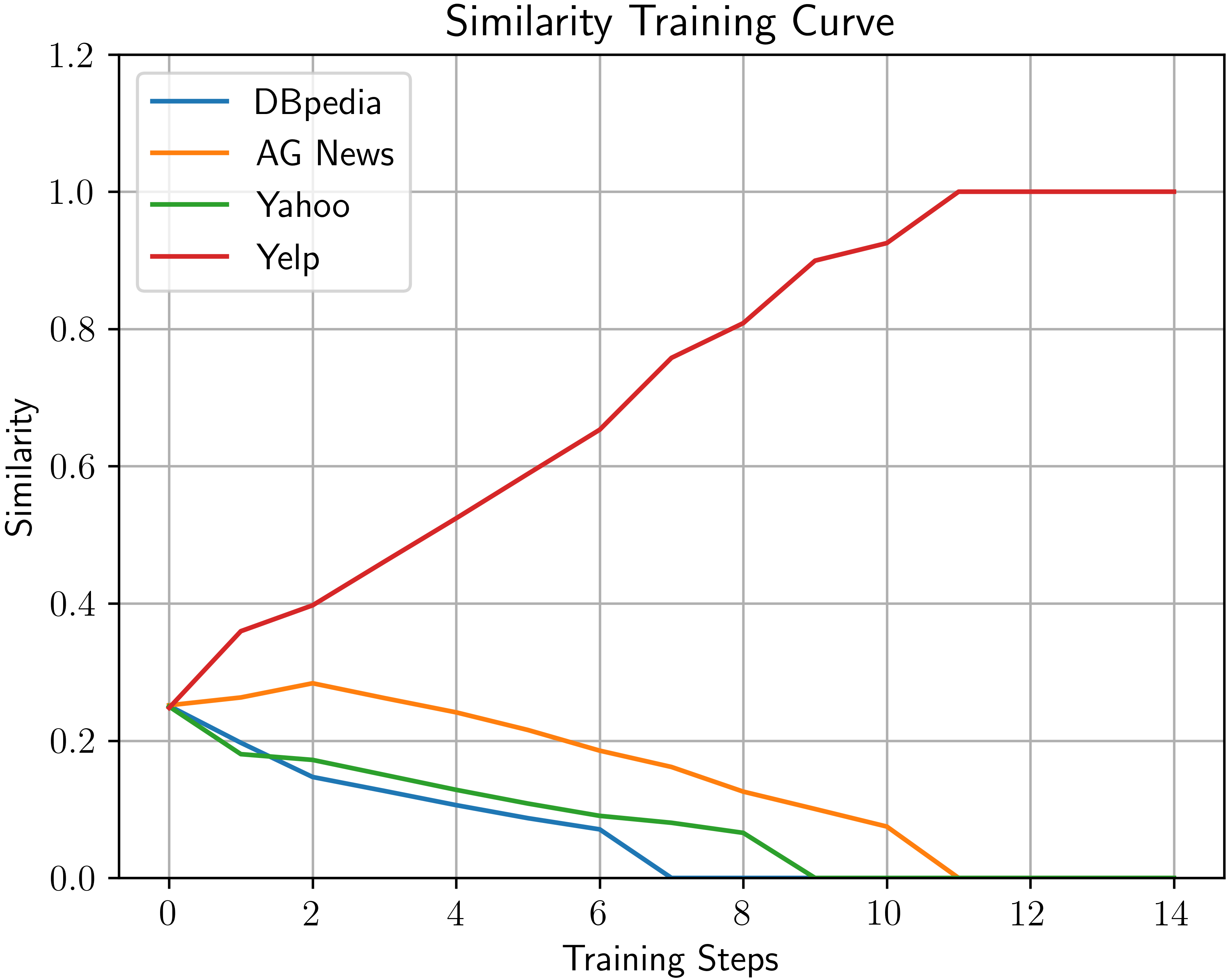}
    \caption{The variation of similarity output by the estimator as training steps increase. We only display a few steps in the early epochs because the similarity does not change afterwards.}
    \label{fig:3}
\end{figure}

We conduct an experiment on datasets from Standard CL Benchmark and show how similarity is learned in the similarity estimator. Figure~\ref{fig:3} illustrates the variation of similarity between the Amazon task and four source tasks during SHLPT training. The similarity is quickly learned within a few steps, and the most similar task, Yelp, also aligns with intuition (both are sentiment analysis tasks for reviews).

\section{Effects of ASC Loss on activation states}
\label{sec:appendix6}
\begin{figure}[htbp]
    \centering
    \includegraphics[width=1.0\linewidth, scale=1.00]{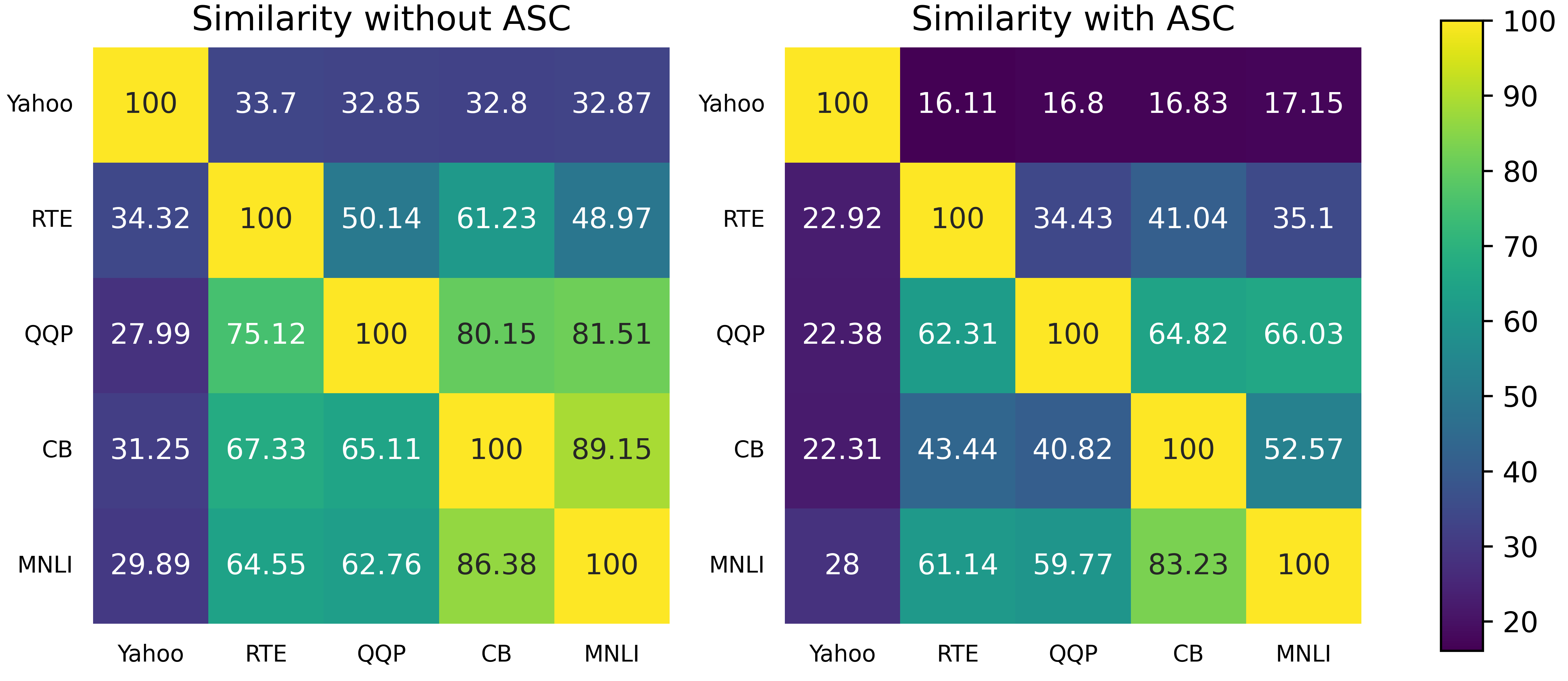}
    \caption{The cosine similarity of activation states at last layer obtained from prompts trained on different tasks.}
    \label{fig:4}
\end{figure}

To investigate how much ASC loss affects activation states, we visualize the cosine similarity of activation states between prompts trained on different tasks (Figure~\ref{fig:4}). The similarity is calculated as follows: we average the similarity values of activation states from different prompts for all samples on each dataset (row). For the figure on the right, we added ASC Loss during the training process to diverge activation states from other tasks. As shown in Figure~\ref{fig:4}, activation states vary across different tasks, and adding ASC Loss reduces the similarity of these tasks' activation states, thereby promoting more diverse neuron activation and a more varied utilization of pre-training knowledge.

\section{Standard Deviations}
\label{sec:appendix7}
Table~\ref{table:app6}, ~\ref{table:app7}, ~\ref{table:app8} report the standard deviations of the results from Table~\ref{table:2} (on Standard CL Benchmark and Large Number of Tasks), ~\ref{table:3} (on Negative Transfer Benchmark), ~\ref{table:5} (ablation studies). Based on the standard deviation results, we find that the performance of SHLPT on the Large Number of Tasks benchmark is significantly better than other baselines. While on the Standard CL Benchmark and Negative Transfer Benchmark, SHLPT also shows a considerable improvement. We believe that the variations in improvement across different benchmarks may stem from this reason: the Large Number of Tasks benchmark contains more datasets, thus resulting in a more significant cumulative performance gain for each task through transfer learning.

\section{Backward Transfer Scores and Forward Transfer Scores}
\label{sec:appendix8}
We further compare SHLPT and the baseline methods in terms of backward transfer scores and forward transfer scores, as defined in \citet{wang2024comprehensive}. Table~\ref{table:app9} reports the average results across three orders/sequences on three benchmarks. 

The prompt-based methods, such as L2P, show no forgetting phenomenon, resulting in a backward transfer score of 0. Conversely, other baselines have negative backward transfer scores, as subsequent tasks induce forgetting, leading to impaired performance in previous tasks.

We use the forward transfer scores to measure negative transfer in lifelong learning, similar to the negative transfer gap in \citet{wang2019characterizing}. A negative forward transfer score indicates negative transfer, while a positive score indicates positive transfer. As shown in Table~\ref{table:app9}, our method SHLPT achieves the best forward transfer score and exhibits the most effective mitigation of negative transfer in three benchmarks.

\begin{table*}[htbp]
    \centering
    \begin{small}
    \begin{tabular}{l|llll}
    \toprule
    Dataset name & Category & Task & Domain & Metric\\
    \midrule
    1. Yelp & CL benchmark & sentiment analysis & Yelp reviews & accuracy  \\
    2. Amazon & CL benchmark & sentiment analysis & Amazon reviews & accuracy  \\
    3. DBpedia & CL benchmark & topic classification & Wikipedia & accuracy  \\
    4. Yahoo & CL benchmark & topic classification & Yahoo Q\&A & accuracy \\
    5. AG News & CL benchmark & topic classification & news & accuracy \\
    6. MNLI & GLUE & NLI & various & accuracy \\
    7. QQP & GLUE & paraphrase & detection Quora & accuracy \\
    8. RTE & GLUE & NLI & news, Wikipedia & accuracy \\
    9. SST2 & GLUE & sentiment analysis & movie reviews & accuracy \\
    10. WiC & SuperGLUE & word sense disambiguation & lexical databases & accuracy  \\
    11. CB & SuperGLUE & NLI & various & accuracy\\
    12. COPA & SuperGLUE & QA & blogs, encyclopedia & accuracy   \\
    13. BoolQ & SuperGLUE & boolean QA & Wikipedia & accuracy   \\
    14. MultiRC & SuperGLUE & QA & various & accuracy \& F1\\
    15. IMDB & Other &  sentiment analysis & movie reviews & accuracy\\
    16. SQuAD v2 & Other & extractive QA & Wikipedia& nF1 \& EM \\
    \bottomrule
    \end{tabular}
    \end{small}
    \caption{Details of 16 tasks used in our experiments. For datasets utilizing two metrics, the primary metric is the one listed first.}
    \label{table:app1}
\end{table*}

\begin{table*}[htbp]
    \centering
    \begin{small}
    \begin{tabular}{c|c}
    \toprule
    Order & Task Sequence   \\
    \midrule
    1 & DBpedia$\rightarrow$ Amazon$\rightarrow$ Yahoo$\rightarrow$ AG News  \\
    2 & DBpedia$\rightarrow$ Amazon$\rightarrow$ AG News$\rightarrow$ Yahoo\\
    3 & Yahoo$\rightarrow$ Amazon$\rightarrow$ AG News$\rightarrow$ DBpedia\\
    \midrule
    \multirow{2}{*}{4} & MNLI$\rightarrow$ CB$\rightarrow$ WiC$\rightarrow$ COPA$\rightarrow$ QQP$\rightarrow$ BoolQ$\rightarrow$ RTE$\rightarrow$ IMDB$\rightarrow$\\
    & Yelp$\rightarrow$ Amazon$\rightarrow$ SST2$\rightarrow$ DBpedia$\rightarrow$ AG News$\rightarrow$ MultiRC$\rightarrow$ Yahoo\\
    \multirow{2}{*}{5} & MultiRC$\rightarrow$ BoolQ$\rightarrow$ WiC$\rightarrow$ MNLI$\rightarrow$ CB$\rightarrow$ COPA$\rightarrow$ QQP$\rightarrow$ RTE$\rightarrow$\\
    & IMDB$\rightarrow$ SST2$\rightarrow$ DBpedia$\rightarrow$ AG News$\rightarrow$ Yelp$\rightarrow$ Amazon$\rightarrow$ Yahoo\\
    \multirow{2}{*}{6} & Yelp$\rightarrow$ Amazon$\rightarrow$ MNLI$\rightarrow$ CB$\rightarrow$ COPA$\rightarrow$ QQP$\rightarrow$ RTE$\rightarrow$ IMDB$\rightarrow$\\
    & SST2$\rightarrow$ DBpedia$\rightarrow$ AG News$\rightarrow$ Yahoo$\rightarrow$ MultiRC$\rightarrow$ BoolQ$\rightarrow$ WiC\\
    
    \bottomrule
    \end{tabular}
    \end{small}
    \caption{Different orders of task sequences used in existing benchmark experiments. Orders 1-3 are employed for Standard CL Benchmark. Orders 4-6 are employed for Large Number of Tasks.}
    \label{table:app2}
\end{table*}

\begin{table*}[htbp]
    \centering 
    \begin{small}
    \begin{tabular}{c|c}
    \toprule
    Sequence ID & Task Sequence \\
    \midrule
    1 & Yahoo$\rightarrow$ RTE$\rightarrow$ QQP$\rightarrow$ CB$\rightarrow$ MNLI\\
    2 & QQP$\rightarrow$ RTE$\rightarrow$ SQuAD v2$\rightarrow$ MNLI$\rightarrow$ CB\\
    3 & MultiRC$\rightarrow$ RTE$\rightarrow$ SQuAD v2$\rightarrow$ WiC$\rightarrow$ MNLI\\ 
    \bottomrule
    \end{tabular}
    \end{small}
    \caption{Three different task sequences used in Negative Transfer Benchmark.}
    \label{table:app3}
    
\end{table*}

\begin{table*}[htbp]
    \centering
    \resizebox{\textwidth}{!}{
    \begin{small}
    \begin{tabular}{ccccccccccccc}
    \toprule
        & WiC & MultiRC & QQP & RTE & CB & MNLI & SQuAD & Yahoo & Yelp & Amazon\\
    \midrule
    Baseline & 62.67 & 52.00 & 86.00 & 78.67 & 87.50 & 88.67 & 65.58 & 74.67 & 58.67 & 54.27\\
    WiC & & \textcolor{red}{55.50} & \textcolor{blue}{84.33} & \textcolor{red}{80.67} & \textcolor{blue}{86.90} & \textcolor{blue}{87.33} & \textcolor{red}{66.74} & \textcolor{red}{75.93} & \textcolor{red}{60.67} & \textcolor{red}{56.27}\\
    MultiRC & \textcolor{blue}{59.33} & & \textcolor{red}{86.33} & \textcolor{blue}{77.00} & \textcolor{red}{87.67} & \textcolor{blue}{88.22} &\textcolor{blue}{64.73} & \textcolor{blue}{74.13} & \textcolor{blue}{58.00} & \textcolor{blue}{45.47}\\
    QQP & 62.67 & \textcolor{red}{53.33} & & \textcolor{blue}{73.67} & 87.50 & \textcolor{red}{88.87} & \textcolor{red}{66.84} & \textcolor{red}{75.27} & \textcolor{blue}{57.87} & \textcolor{blue}{53.73}\\
    RTE & \textcolor{blue}{59.00} & \textcolor{red}{55.67} & 86.00 & &\textcolor{blue}{86.91} & \textcolor{blue}{87.33} & \textcolor{blue}{65.55} & \textcolor{red}{76.13} & \textcolor{red}{60.13} & \textcolor{red}{54.93}\\
    CB & \textcolor{red}{65.67} & \textcolor{blue}{51.00} & \textcolor{red}{89.00} & 78.67 &  & \textcolor{blue}{88.22} & \textcolor{red}{67.94} & \textcolor{red}{75.13} & \textcolor{red}{59.60} & \textcolor{red}{55.47}\\
    MNLI & \textcolor{blue}{59.67} & \textcolor{blue}{51.67} & \textcolor{red}{87.67} & \textcolor{blue}{75.67} & \textcolor{blue}{83.93} & & \textcolor{red}{66.40} &\textcolor{blue}{74.40} & \textcolor{red}{60.13} & \textcolor{red}{54.80}\\
    SQuAD v2 & 62.67 & \textcolor{blue}{48.00} & \textcolor{red}{87.00} & \textcolor{red}{81.00} & \textcolor{blue}{83.93} & \textcolor{blue}{86.44} & & \textcolor{red}{75.07} & \textcolor{red}{58.93} & \textcolor{red}{55.47} \\    
    Yahoo & \textcolor{red}{65.67} & \textcolor{red}{52.33} & \textcolor{blue}{81.00} & \textcolor{blue}{73.33} & \textcolor{blue}{84.52} & \textcolor{blue}{88.00} & \textcolor{red}{66.85} & & \textcolor{red}{60.00} & \textcolor{red}{56.80}\\
    Yelp & \textcolor{blue}{61.33} & \textcolor{red}{54.67} & \textcolor{blue}{81.00} & \textcolor{red}{80.67} & \textcolor{red}{91.67} & \textcolor{blue}{88.45}   & \textcolor{red}{68.23} & \textcolor{red}{76.07} & & \textcolor{red}{55.60}\\
    Amazon & 62.67 & \textcolor{red}{55.00} & \textcolor{blue}{80.67} & \textcolor{blue}{77.00} & \textcolor{blue}{86.90} & \textcolor{blue}{88.44}  & \textcolor{blue}{64.57} & 74.67 & \textcolor{red}{61.07} & \\
    \bottomrule
    \end{tabular}
    \end{small}
    }
    \caption{Each cell in the columns represents the performance of the target task transferred from a specific source task (row). Baseline refers to the task accuracy without transfer. Positive transfers are shown in red while negative transfers are shown in blue.}
    \label{table:app4}
\end{table*}

\begin{table*}[htbp]
    \centering 
    \begin{small}
    \begin{tabular}{c|ccccccc}
    \toprule
    Threshold $\delta$ & 0.02 & 0.04 & 0.06 & 0.08 & 0.10 & 0.15 & 0.20\\
    \midrule
    Accuracy (\%) & $80.00\pm0.20$ & $80.03\pm0.24$ & $\mathbf{80.21\pm0.37}$ & $80.02\pm0.18$ & $80.02\pm0.18$ & $79.57\pm0.25$ & $79.63\pm0.14$ \\
    \bottomrule
    \end{tabular}
    \end{small}
    \caption{Searching results for SHLPT's optimal similarity threshold on the Standard CL Benchmark in Order1. We report the average accuracy after learning the last task.}
    \label{table:app5}
    
\end{table*}

\begin{table*}[htbp]
\centering
\begin{small}
\begin{tabular}{@{}l|cccc|cccc@{}}
\toprule
\multicolumn{1}{l|}{\multirow{2}{*}{Method}} & \multicolumn{4}{c|}{Standard CL Benchmark}                                                                       & \multicolumn{4}{c}{Large Number of Tasks}                                                                       \\ \cmidrule(l){2-9} 
\multicolumn{1}{c|}{}                        & \multicolumn{1}{c}{Order1} & \multicolumn{1}{c}{Order2} & \multicolumn{1}{c|}{Order3} & \multicolumn{1}{c|}{Avg} & \multicolumn{1}{c}{Order4} & \multicolumn{1}{c}{Order5} & \multicolumn{1}{c|}{Order6} & \multicolumn{1}{c}{Avg} \\ \midrule
Finetune & $\pm1.38$ & $\pm0.24$ & \multicolumn{1}{c|}{$\pm3.61$}& $\pm0.67$ & $\pm3.98$ & $\pm0.41$ &\multicolumn{1}{c|}{$\pm0.41$} & $\pm1.45$   \\
Online EWC & $\pm4.37$ & $\pm4.04$& \multicolumn{1}{c|}{$\pm4.16$} & $\pm2.47$ & $\pm9.44$& $\pm6.36$ & \multicolumn{1}{c|}{$\pm5.74$} & $\pm1.69$ \\
ER & $\pm2.47$& $\pm1.04$& \multicolumn{1}{c|}{$\pm3.79$}&$\pm2.27$ & $\pm4.22$& $\pm6.06$& \multicolumn{1}{c|}{$\pm1.06$} & $\pm1.82$\\
Per-task Prompts & $\pm2.20$ & $\pm2.20$ & \multicolumn{1}{c|}{$\pm2.20$} & $\pm2.20$ & $\pm0.92$ & $\pm0.92$ & \multicolumn{1}{c|}{$\pm0.92$} & $\pm0.92$ \\
L2P & $\pm0.98$ & $\pm0.54$ & \multicolumn{1}{c|}{$\pm0.44$} & $\pm0.20$ & $\pm1.00$ & $\pm0.79$ & \multicolumn{1}{c|}{$\pm1.00$} & $\pm0.34$   \\
CODA-Prompt & $\pm2.17$ & $\pm1.40$ & \multicolumn{1}{c|}{$\pm5.18$}& $\pm1.72$ & $\pm0.84$ & $\pm0.01$ & \multicolumn{1}{c|}{$\pm0.35$} & $\pm0.33$ \\
ProgPrompt & $\pm3.39$ & $\pm2.19$ & \multicolumn{1}{c|}{$\pm0.49$}& $\pm1.86$& $\pm1.19$& $\pm1.50$ &\multicolumn{1}{c|}{$\pm0.68$} & $\pm0.41$  \\
SHLPT(ours) & $\pm0.37$ & $\pm0.40$ & \multicolumn{1}{c|}{$\pm0.47$}& $\pm0.07$ & $\pm0.39$ & $\pm1.32$ & \multicolumn{1}{c|}{$\pm0.45$} & $\pm0.42$\\
\bottomrule
\end{tabular}
\end{small}
\caption{Standard deviations of the related metrics of SHLPT and baseline methods on Standard CL Benchmark and Large Number of Tasks.}
\label{table:app6}
\end{table*}

\begin{table*}[t]
\centering
\begin{small}
\begin{tabular}{@{}l|cccc@{}}
\toprule
\multicolumn{1}{l|}{\multirow{2}{*}{Method}} & \multicolumn{4}{c}{Negative Transfer Benchmark}\\ 
\cmidrule(l){2-5} 
\multicolumn{1}{c|}{} & \multicolumn{1}{c}{Seq1} & \multicolumn{1}{c}{Seq2} & \multicolumn{1}{c|}{Seq3} & \multicolumn{1}{c}{Avg} \\ 
\midrule
Finetune & $\pm4.69$ & $\pm1.13$ & \multicolumn{1}{c|}{$\pm9.44$}& $\pm1.67$ \\
Online EWC & $\pm3.67$ & $\pm7.16$ &\multicolumn{1}{c|}{$\pm6.06$} & $\pm5.38$ \\
ER & $\pm8.42$ & $\pm5.37$ & \multicolumn{1}{c|}{$\pm1.23$}& $\pm1.92$ \\
Per-task Prompts & $\pm0.77$& $\pm1.36$ & \multicolumn{1}{c|}{$\pm1.73$}& $\pm1.25$ \\
L2P & $\pm1.40$& $\pm1.79$ & \multicolumn{1}{c|}{$\pm1.31$}& $\pm1.46$ \\
CODA-Prompt & $\pm0.15$ & $\pm0.78$ & \multicolumn{1}{c|}{$\pm2.26$} & $\pm0.67$\\
ProgPrompt & $\pm1.25$ & $\pm1.13$ & \multicolumn{1}{c|}{$\pm2.44$} & $\pm1.39$\\
SHLPT(ours) & $\pm1.21$ & $\pm0.78$ & \multicolumn{1}{c|}{$\pm0.39$}& $\pm0.28$\\
\bottomrule
\end{tabular}
\end{small}
\caption{Standard deviations of the related metrics of SHLPT and baseline methods on Negative Transfer Benchmark.}
\vspace{-1.2em}
\label{table:app7}
\end{table*}

\begin{table*}
    \centering
    \resizebox{\columnwidth}{!}{
    \begin{small}
    \begin{tabular}{l|cccc|c}
    \toprule
     Model& Seq1 & Seq2 & Seq3 & Seq4 & Avg \\
    \midrule
    -ASE & $\pm1.73$ & $\pm0.29$ & $\pm1.07$ & $\pm1.68$ & $\pm0.75$\\
    -ASC & $\pm0.62$ & $\pm0.36$ & $\pm1.57$ & $\pm0.75$ & $\pm0.20$\\
    -HSC & $\pm0.31$ & $\pm1.39$ & $\pm0.39$ & $\pm0.87$ & $\pm0.33$\\
    -STT & $\pm1.21$ & $\pm0.78$ & $\pm0.39$ & $\pm0.43$ & $\pm0.11$\\
    -ASC,-HSC & $\pm1.36$ & $\pm0.49$ & $\pm0.80$ & $\pm0.71$ & $\pm0.52$\\
    \midrule
    \textbf{SHLPT}& $\pm1.21$ & $\pm0.78$ & $\pm0.39$ & $\pm0.37$ & $\pm0.29$\\
    \bottomrule
    \end{tabular}
    \end{small}
    }
    \caption{Standard deviations of the related metrics of SHLPT and the ablations.}
    \label{table:app8}
\end{table*}

\begin{table*}[t]
\centering
\begin{small}
\begin{tabular}{@{}l|cccccc@{}}
\toprule
\multicolumn{1}{l|}{\multirow{2}{*}{Method}} & \multicolumn{2}{c|}{Standard CL Benchmark} & \multicolumn{2}{c|}{Large Number of Tasks} & \multicolumn{2}{c|}{Negative Transfer Benchmark} \\ 
\cmidrule(l){2-7} 
\multicolumn{1}{c|}{} & \multicolumn{1}{c}{BWT} & \multicolumn{1}{c|}{FWT} & \multicolumn{1}{c}{BWT} & \multicolumn{1}{c|}{FWT}  & \multicolumn{1}{c}{BWT} &\multicolumn{1}{c|}{FWT}\\
\midrule
Finetune & -62.00 & \multicolumn{1}{c|}{-0.69} & -64.41 & \multicolumn{1}{c|}{-3.73} & -44.02 & \multicolumn{1}{c|}{-3.43} \\
Online EWC & -21.56 & \multicolumn{1}{c|}{-2.17} & -24.27 & \multicolumn{1}{c|}{-4.98} & -13.30 & \multicolumn{1}{c|}{-3.94}\\
ER & -15.82 & \multicolumn{1}{c|}{-1.21} & -16.62 & \multicolumn{1}{c|}{-4.55} & -9.04 & \multicolumn{1}{c|}{-8.06}\\
L2P & 0.00 & \multicolumn{1}{c|}{-4.34} & 0.00 & \multicolumn{1}{c|}{-2.30} & 0.00 & \multicolumn{1}{c|}{-0.85}\\
CODA-Prompt & 0.00 & \multicolumn{1}{c|}{-1.80} & 0.00 & \multicolumn{1}{c|}{0.56} & 0.00 & \multicolumn{1}{c|}{-0.49}\\
ProgPrompt & 0.00 & \multicolumn{1}{c|}{-3.57} &  0.00 & \multicolumn{1}{c|}{-3.88} & 0.00 & \multicolumn{1}{c|}{-3.17}\\
SHLPT(ours) & 0.00 & \multicolumn{1}{c|}{\textbf{1.36}} &  0.00 & \multicolumn{1}{c|}{\textbf{1.45}} & 0.00 & \multicolumn{1}{c|}{\textbf{0.99}}\\
\bottomrule
\end{tabular}
\end{small}
\caption{The average backward transfer scores (BWT) and forward transfer scores (FWT) on Standard CL Benchmark, Large Number of Tasks and Negative Transfer Benchmark.}
\vspace{-1.2em}
\label{table:app9}
\end{table*}

\end{document}